**Daria D. Guseva**[1]
**Olga A. Mitrofanova**[2]

# Automatic keyphrase extraction and annotation: modern theoretical approaches and practical solutions for text and speech

[1] Saint-Petersburg State University,
7/9 Universitetskaya Embankment, Saint Petersburg, 199034, Russia
*E-mail:* daria.guseva@spbu.ru
ORCID: 0009-0005-1753-9904

[2] Saint-Petersburg State University,
7/9 Universitetskaya Embankment, Saint Petersburg, 199034, Russia
*E-mail:* o.mitrofanova@spbu.ru
ORCID: 0000-0002-3008-5514

**Abstract:** The exponential growth of textual and audiovisual information has made the task of automatic keyphrase extraction (KE) increasingly significant. This article provides a comprehensive analysis of contemporary theoretical approaches and practical solutions for KE across both text and speech modalities. The primary contribution of this work is its systematic synthesis of these often-disparate research strands into a unified analytical framework, highlighting the evolution of the field from statistical methods towards large language models (LLMs) and end-to-end speech processing. We examine the stages of KE, the characteristics of keyphrases in written and spoken language, and terminological nuances. Various methods for automatic KE are discussed and analyzed in detail: statistical, hybrid, machine learning-based, and structural. The review dedicates substantial attention to emerging paradigms, including keyphrase generation using LLMs, and provides a detailed overview of methodologies and challenges in automatic corpus annotation. Furthermore, we specifically analyze current directions and inherent difficulties in KE for spoken language, comparing transcript-based and end-to-end acoustic approaches. This synthesis leads us to conclude that the field is moving towards a more integrated, context-aware paradigm. Future progress will depend on addressing key challenges such as data scarcity for low-resource languages, effective multimodal fusion, and the nuanced evaluation of generative KE systems.

**Keywords:** Automatic keyphrase extraction; Spoken language processing; Speech summarization; Automatic annotation; Computational linguistics; Corpus linguistics

**Acknowledgements:** This research was supported by Saint-Petersburg State University, project № 123042000068-8

**How to cite:** Guseva, D. D., Mitrofanova, O. A. (2026). Automatic keyphrase extraction and annotation: modern theoretical approaches and practical solutions for text and speech, *Research Result. Theoretical and Applied Linguistics*, 12 (1), 90–122. DOI: 10.18413/2313-8912-2026-12-1-0-4

**УДК 81’32, 81’33** **DOI: 10.18413/2313-8912-2026-12-1-0-4**

**Гусева Д. Д.**[1]
**Митрофанова О. А.**[2]

# Автоматическое выделение ключевых выражений и аннотирование: современные теоретические подходы и практические решения для работы с текстом и речью

[1] Санкт-Петербургский государственный университет
Университетская набережная, 7/9, Санкт-Петербург, 199034, Россия
*E-mail:* daria.guseva@spbu.ru
ORCID: 0009-0005-1753-9904

[2] Санкт-Петербургский государственный университет
Университетская набережная, 7/9, Санкт-Петербург, 199034, Россия
*E-mail:* o.mitrofanova@spbu.ru
ORCID: 0000-0002-3008-5514

**Аннотация:** Экспоненциальный рост объёмов текстовой и аудиовизуальной информации значительно повышает значимость задачи автоматического выделения ключевых выражений. В данной статье представлен всесторонний анализ современных теоретических подходов и практических решений для автоматического выделения ключевых выражений как в текстовой, так и в речевой модальностях. Основной вклад работы заключается в систематическом синтезе этих зачастую разрозненных направлений исследований в рамках единой аналитической структуры, демонстрирующей эволюцию области от статистических методов к большим языковым моделям и сквозной обработке речи. Мы рассматриваем этапы автоматического выделения ключевых выражений, характеристики ключевых выражений в письменной и устной речи, а также терминологические нюансы. Подробно обсуждаются и анализируются различные методы автоматического выделения ключевых выражений: статистические, гибридные, основанные на машинном обучении и структурные. В обзоре значительное внимание уделено новым парадигмам, включая генерацию ключевых выражений с использованием больших языковых моделей, а также представлен детальный обзор методологий и проблем автоматической аннотации корпусов. Кроме того, мы анализируем текущие направления и трудности, присущие задаче автоматического выделения ключевых выражений из устной речи, и сравниваем подходы, основанные на расшифровке и сквозном акустическом анализе. Этот синтез позволяет сделать вывод о переходе области к более интегрированной, контекстно-ориентированной парадигме. Дальнейший прогресс будет зависеть от решения ключевых задач, таких как нехватка данных для малоресурсных языков, эффективное мультимодальное объединение и тонкая оценка генеративных систем автоматического выделения ключевых выражений.

**Ключевые слова:** Автоматическое выделение ключевых выражений; Обработка устной речи; Суммаризация речи; Автоматическое аннотирование; Компьютерная лингвистика; Корпусная лингвистика

**Благодарности:** Исследование выполнено при поддержке СПбГУ, шифр проекта 123042000068-8

**Информация для цитирования:** Гусева Д. Д., Митрофанова О. А. Автоматическое выделение ключевых выражений: современные теоретические подходы и практические решения // Научный результат. Вопросы теоретической и прикладной лингвистики. 2026. Т. 12. № 1. С. 90–122. DOI: 10.18413/2313-8912-2026-12-1-0-4

## 1. Introduction

The task of automatic keyphrase extraction (KE) is gaining increasing relevance amidst the exponential growth in both textual and audiovisual information. Effective keyphrase extraction methods enable the structuring and analysis of large volumes of data, automating processes of search, classification, and summarization. This has broad applications across various fields, including computational and cognitive linguistics, psycholinguistics, philology, communication theory, computer science, economics, and more. The selection of extraction methods, in this context, is considered a complex and debated issue. Within the domain of text linguistics, this task is regarded as one of the most challenging (Papusha, 2008). Areas where KE is essential include natural language processing, information retrieval, machine learning (ML) and artificial intelligence (AI), education, databases, marketing and advertising, business analytics, translation, and social media analysis.

The significance of this field is further evidenced by the substantial and growing body of research. A search of the Dimensions[1] engine reveals 24,338 publications on keyphrase processing and applications over the past 50 years (see Fig. 1). The number of papers devoted to keyphrase extraction, assignment, search, etc. has increased exponentially since the 2000s, reaching a maximum of 1137 items in 2023. Distribution of research papers over fields of knowledge indicates the dominance of three domains: Biomedical and Clinical Sciences (12,485 papers), Information and Computing Sciences (9,836 papers), Language, Communication and Culture (974 papers), cf. Figure 2.

VOS Viewer data visualization[2] allows users to navigate in co-authorship and citation networks. For top 500 researchers we registered 531 co-authorship links in 2739 total co-authorships distributed over 200 clusters, cf. Fig. 3. VOS Viewer indicates high citation density and allows to point cross-reference network with fundamental publications on the topic at its core, cf. Fig. 4. For top 500 researchers the network includes 8994 citation links, 17593 total citations distributed over 52 clusters. However, the problems of keyword annotation and spoken text processing are not sufficiently covered in publications, so links to them are not included in the top results of scientometric tools. Our work aims to restore the interest of the academic community in these aspects.

[1] Available at: https://www.dimensions.ai/ (Accessed 22.11.2025)

[2] Available at: https://app.vosviewer.com/ (Accessed 22.11.2025)

**Figure 1.** Publication activity: papers on keyphrase analysis and processing via Dimensions over 50 years
**Рисунок 1.** Публикационная активность: статьи по анализу и обработке ключевых выражений в Dimensions за последние 50 лет

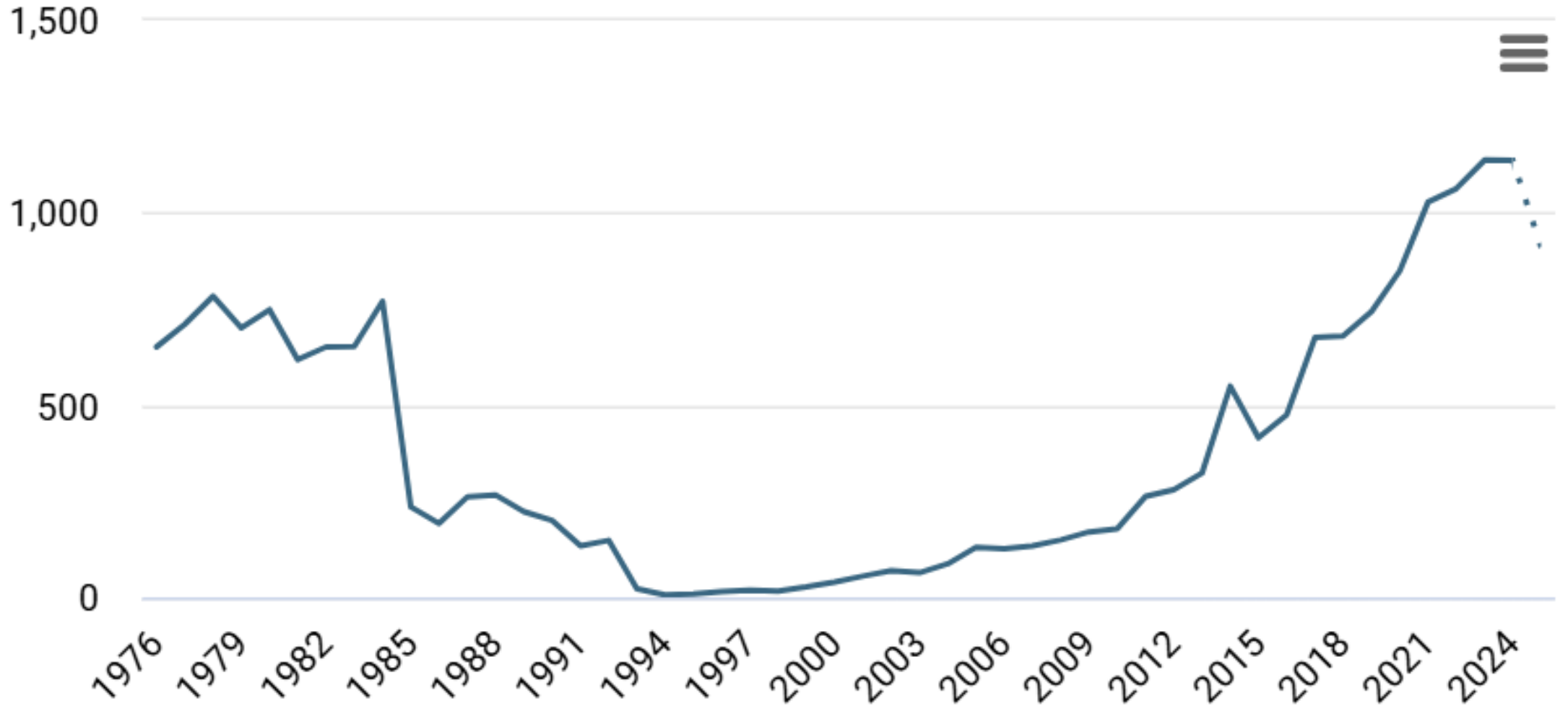

**Figure 2.** Distribution of research papers on keyphrase analysis and processing over fields of knowledge (top 10 domains)
**Рисунок 2.** Распределение научных работ по анализу и обработке ключевых выражений по областям знаний (топ-10 областей)

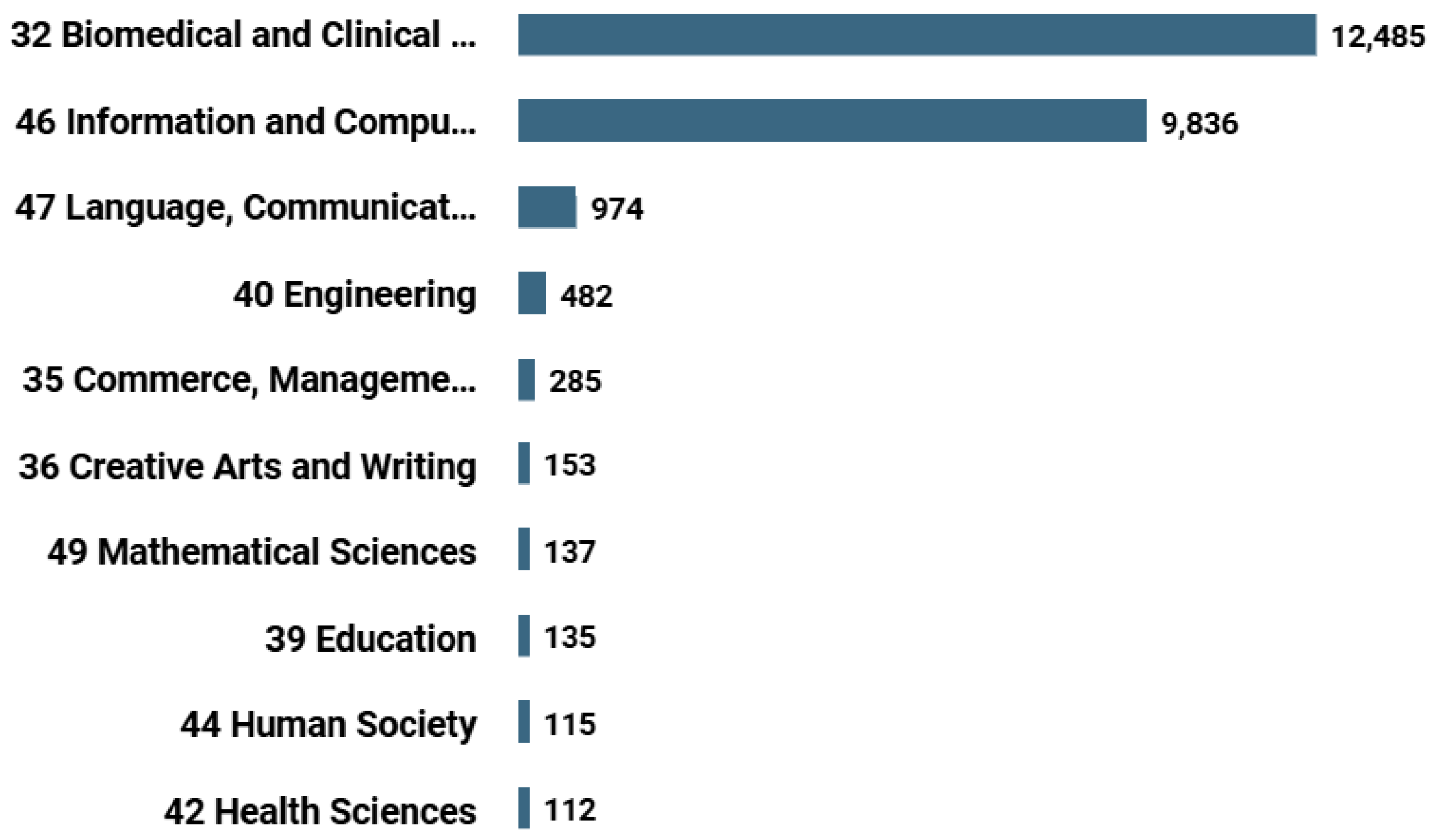

**Figure 3.** VOSViewer co-authorship network in papers on keyphrase analysis and processing
**Рисунок 3.** Сеть соавторов статей по анализу и обработке ключевых выражений в VOSViewer

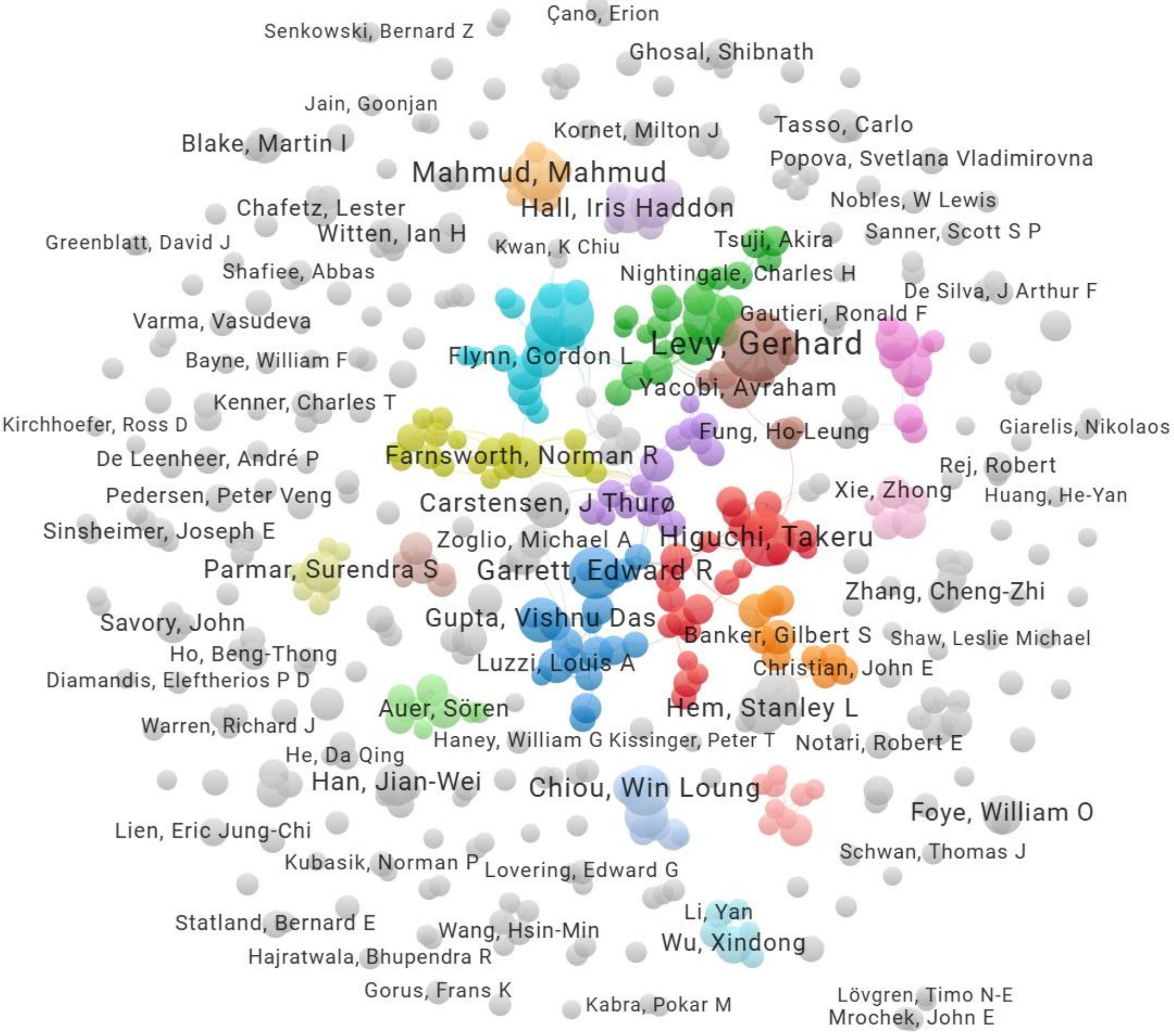

**Figure 4.** VOSViewer network of citations in papers on keyphrase analysis and processing
**Рисунок 4.** Сеть цитирований в статьях по анализу и обработке ключевых выражений в VOSViewer

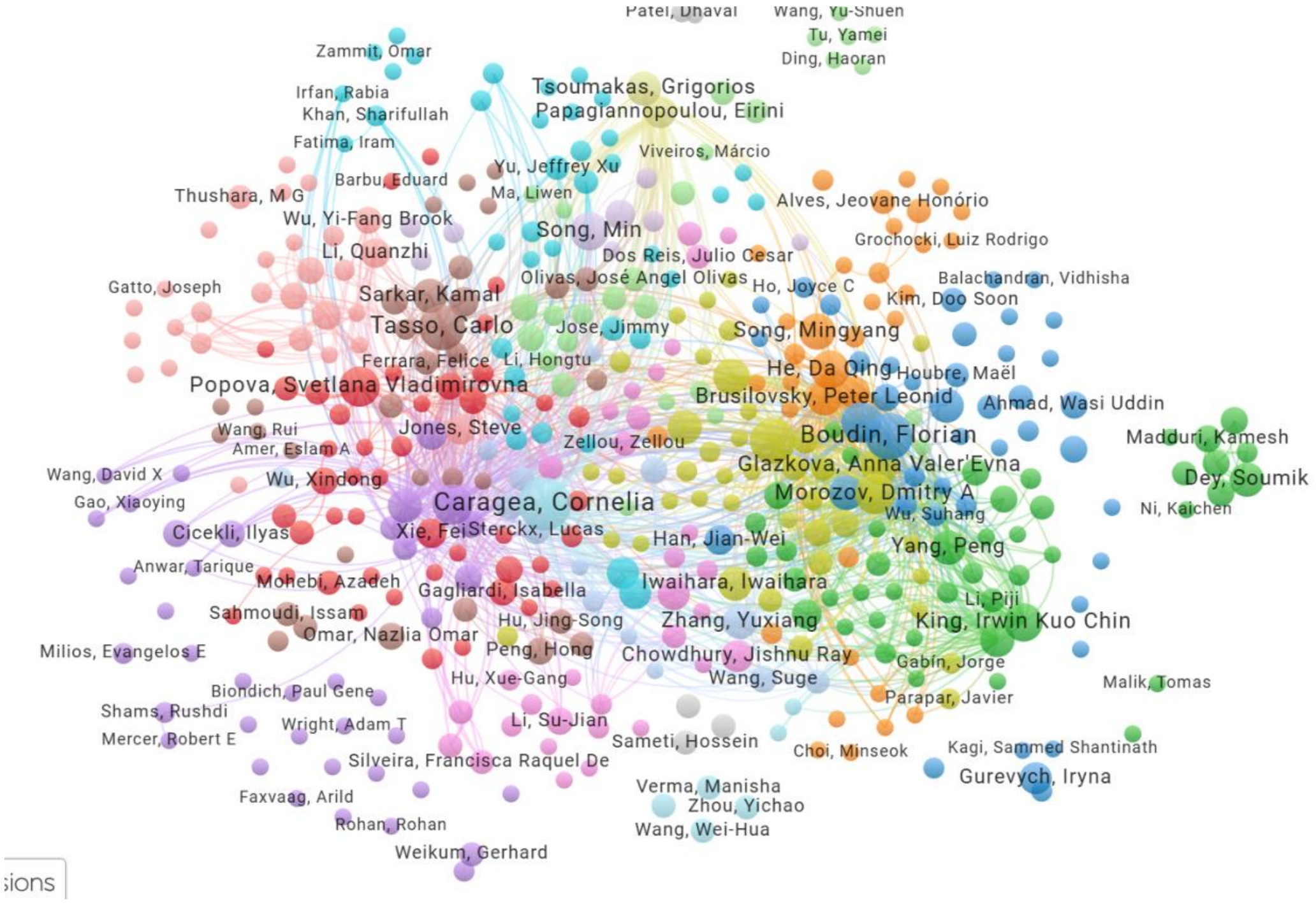

Different scientific disciplines exhibit significant divergences in their approaches to studying and defining keyphrases. For instance, information retrieval focuses on the practical application of keyphrases for information extraction, while psycholinguistics concentrates on their cognitive and linguistic properties. The issue of approach unification is even proposed to be addressed through musicological analysis, based on common patterns in the semantic structure of poetic texts and musical compositions (Antipina and Prokhorenkova, 2020). The multiplicity of approaches leads to variations in terminology and methodology.

This article examines contemporary theoretical approaches and practical solutions for the task of automatic keyphrase extraction. Within this review, we analyze the stages of automatic KE, the features characterizing them in written and spoken language, and the terminological specifics that have emerged across different scientific domains. We also discuss various methods of automatic KE, including statistical, hybrid, machine learning-based, and structural approaches, and analyze their advantages and limitations. Particular attention is paid to the specifics of KEs in spoken language, including their acoustic characteristics and their impact on perception, as the extraction of these units from speech presents unique challenges and remains significantly underexplored compared to written text analysis. Furthermore, an overview of current trends in automatic KE from spoken language is presented, encompassing the use of deep learning methods and automatic annotation. The analysis of these methods will allow us to evaluate the effectiveness of different approaches and identify promising directions for future research.

This study is framed by the following research questions aimed at advancing our understanding of keyphrase extraction:

RQ1: How do different research disciplines (information retrieval, psycholinguistics, speech processing, etc.) define and operationalize keyphrases in their methodologies?

RQ2: How do these methodologies contribute to keyphrase extraction procedures across disciplines?

RQ3: What opportunities arise from combining psycholinguistic theories with information retrieval metrics for improved keyphrase extraction?

RQ4: What are limitations of current automatic keyphrase extraction methods as regards text genres, domains, written or spoken texts?

RQ5: What are the most significant language-specific challenges in keyphrase processing?

RQ6: How LLMs may improve automatic keyphrase extraction and generation?

RQ7: What are the perspectives of integration of multimodal information into keyphrase processing?

Our review is based on the following methodological principles: in choosing theoretical foundations, we focus on those approaches that have been implemented or are applicable to Russian-language data; in choosing datasets, corpora and models we take into account their accessibility to the scientific community, in choosing technical solutions we consider their scalability and reproducibility.

## 2. Theoretical foundations of keyphrase extraction

### 2.1. Terminology and Definitions

The terms “keyphrase,” (“ключевое выражение/словосочетание”) and “keyword,” (“ключевое слово”) can often be used synonymously in most studies, including those mentioned in the next paragraph. In this context, “keyword” frequently refers to both single words and multi-word expressions, driven by the necessity of using collocations rather than isolated words for a more accurate and precise reflection of content (Bolshakova et al., 2011).

Additionally, the Russian-language literature includes terms such as “ключевые термины” (“key terms”) (Grineva and Grinev, 2009), “слова-концепты” (“concept-words”), “слова-лейтмотивы” (“leitmotif-words”) (Svetozarova and Shtern, 1989), “лексические доминанты” (“lexical dominants”), “семантические доминанты” (“semantic dominants”), and “смысловые вехи” (“semantic milestones”) (Shekhtman, 2005). In English-language literature, “keyword,” “keyphrase,” and “key item” are used (Wilson, 2013). The concept of “keyness” is also employed in English literature, referring to a statistical measure that reflects the most content-rich words in a text. Its analysis is used to identify elements that occur statistically significantly more or less frequently in one corpus compared to another, and can focus on differences (keyness-D) or similarities (keyness-S) between corpora (Kilgarriff, 2009; Gabrielatos, 2018; Evert, 2022).

The variety of terminology used to identify keyphrases is illustrated in Figure 5.

**Figure 5.** Terminological landscape of keyphrases
**Рисунок 5.** Терминологическое разнообразие для обозначения ключевых выражений

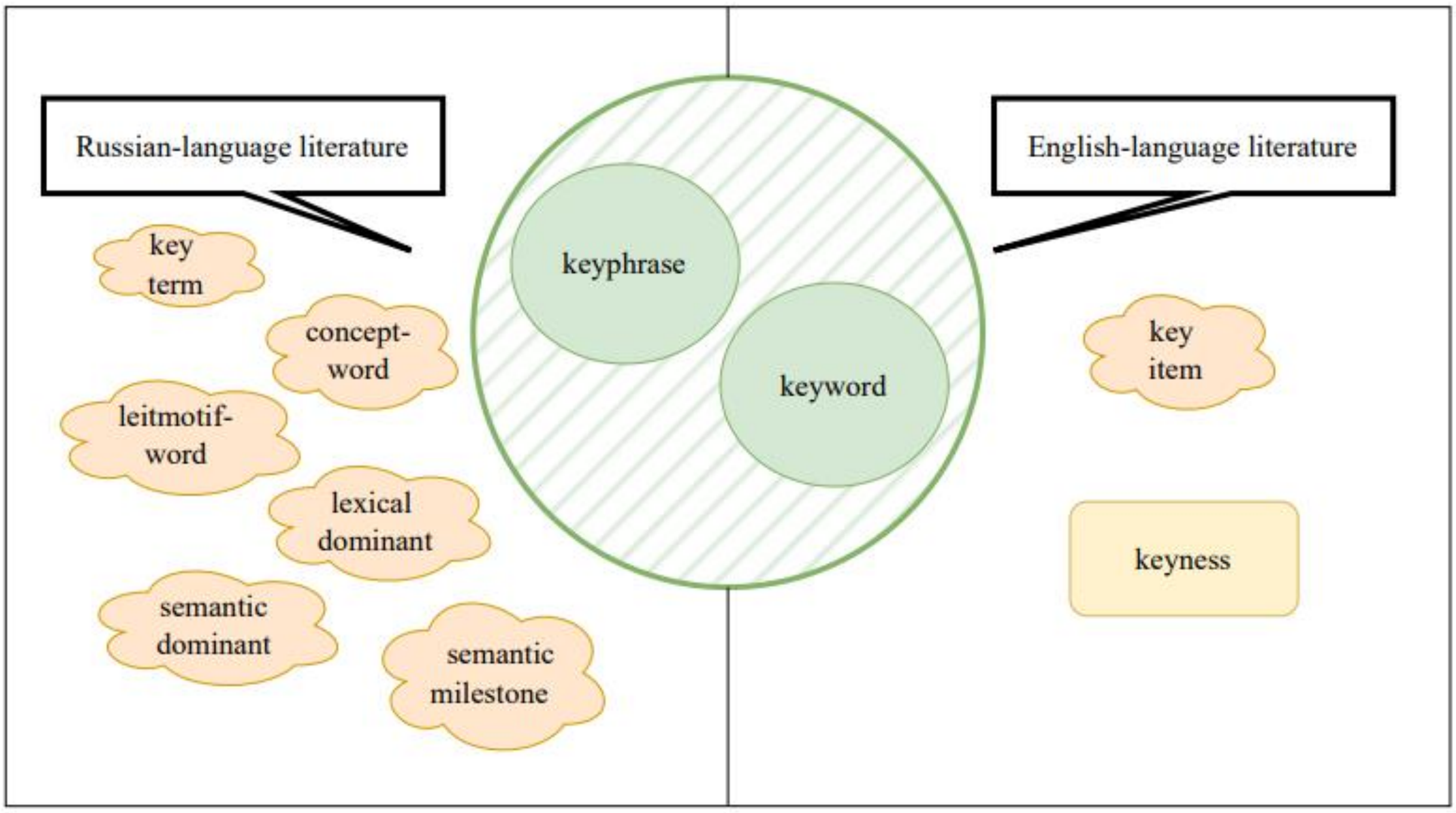

In this work, we will use the term “keyphrase,” which, in our view, leads to fewer ambiguities. When citing the work of other researchers, their original terminology may be preserved.

We proceed to an examination of the core definitions of keyphrases presented in the Russian-language domain literature. In text perception studies, a psycholinguistic approach is often adopted, where a set of keywords is considered a condensed version of the original, expanded text (“текст-примитив”), possessing wholeness and coherence (Sakharny and Shtern, 1988). Keywords convey the text’s theme, and their ordering refects the rheme (Sakharny, 1982). It is believed that any written or spoken text can be compressed into a set of keywords. Consequently, the content structure, defined by the distribution of keyphrases within a text, can be described as semantically significant, as it enables one to understand the main meaning of the text. This is particularly important for text perception. Keyphrases extracted from a text form its linguistic image, which can be linearly expanded back to reconstruct the content of the compressed text.

The foundational research conducted by Sakharny and Shtern (1988) warrants particular scholarly attention within the context of this review. Their seminal investigation remains highly relevant as a cornerstone for contemporary KE studies, and crucially, it constitutes one of the relatively few methodologies explicitly built upon the empirical analysis of keyphrases functioning within spoken discourse.

The work by Sakharny and Shtern (1988) presents a review of a series of experiments using various texts (fiction, scientific and technical, spontaneous dialogues, etc.), which were used to extract the keywords, reconstruct text from the sets of keywords, and reconstruct spoken text presented against a background of white noise. In the latter case, experiments were conducted both with and without prior familiarization of participants with the sets of keywords. The researchers investigated patterns of text compression into the sets of keywords, their structure, their connection to the original text's structure, and the influence of the keywords on the noise resistance of the original text. Typically, words from the initial part of the text are prioritized for the set of keywords, with further expansion occurring through refinement of the characteristics of the selected keyword(s). Nouns (in their base form) and/or noun-adjective combinations denoting the themes of text fragments often serve this role. From a syntactic perspective, a chain-like structure is characteristic of the set of keywords (where the preceding rheme becomes the theme in the subsequent fragment). Disruptions arise when the original text has a more complex structure. Experiments also demonstrated that prior familiarization of subjects with the set of keywords can improve the perception of the original text under both external and internal interference (white noise, poor language proficiency, missed words in the text). This confirms that keywords create contextual connections and associations that facilitate text comprehension, and the mechanism underlying improved perception after familiarization with the set of keywords is universal, operating for both Russian speakers in noisy conditions and foreign learners of Russian.

In scientific texts, a set of keyphrases is defined as an invariant of content, reflecting the second stage of scientific text generation and perception (after the title, before the abstract and full text) (Kamshilova, 2013). Keyphrases are found at all these stages but with varying quantitative and substantive content, and they are typically present in strong positions within the text (first sentences and paragraphs).

Since keyphrases indicate "the subject area of knowledge, space, and time, equally understood by members of the same professional community, facilitating dialogue between them" (Moskvitina, 2009, p. 279), they can attract attention to texts, which aligns with the goals of information retrieval. In this context, the set of keyphrases includes the most significant units of a text and serves to form the search profile of documents (indexing), allowing for their concise characterization and retrieval (Jacquemin and Bourigault, 2003; Dostal, 2011). The applicability of keywords as descriptive metadata is universal, extending across diverse information objects, such as text, image, audio file, video file, and digital resources, disseminated online, thereby establishing their relevance for core information retrieval tasks (Abramov, 2011). Their use enhances search efficiency, creates compact indexes, improves ranking, and optimizes user queries. In a general sense, the approach to keyphrase extraction from an information retrieval perspective aims to create content that maximally matches user queries and increases website visibility in search results.

Internet texts are presented as HTML pages containing information about the website's content. Search engines index these pages, using keyphrase extraction to rank search results. Most information retrieval systems are based on a vector data model, where text is represented as a vector (set) of keyphrases. Search engines compare user query vectors with the vectors of indexed web pages. Web pages containing keyphrases that match the query will have more similar vectors and, consequently, will be considered more relevant (Bolshakova et al., 2011).

Keyphrases are also defined as an "information portrait" of a topic, obtained as a result of searching for a specific query (Bolshakova et al., 2011). Their relevance and pertinence are determined by statistical and semantic algorithms. Evaluation criteria can include features such as the frequency of occurrence, density, and distribution of keyphrases within a document, as well as the use of semantic indexing algorithms (specifically, LSI, Latent Semantic Indexing), i.e., the presence of keyphrases semantically related to the target keyphrases. The final portrait can be presented, for example, as a "semantic map" or a table demonstrating search results (Bolshakova et al., 2011).

From the perspective of semantics, a branch of linguistics, keyphrases are defined as "семантические доминанты" ("semantic dominants") (Shekhtman, 2005), which provide access to the hidden meaning of a text, i.e., they are considered in conjunction with the situational contexts they represent. With this approach, keyphrase types are identified at three levels of semantic structure organization: at the content level – "опорные слова" ("supporting words,") at the meaning level – "смысловые вехи" ("semantic milestones,") and at the intent level – ("слова-концепты") ("concept-words.") Based on this, it can be stated that keyphrases convey the meaning of a text at all levels of its comprehension (Moskvitina, 2009).

**2.2. Features of Keyphrases**

Traditionally, the primary features of keyphrases are considered to include their presence in high-frequency vocabulary; semantic coherence; and part-of-speech belonging to nouns, verbs, and adjectives (Vanyushkin and Grashchenko, 2016). However, the identification of these features is quite conventional. In reality, keyphrases are not always frequent, may not be semantically connected (e.g., in conversational text characterized by rapid topic shifts), and can belong to other parts of speech.

The features characterizing keyphrases can also depend on the functional style of the text. For instance, in literary texts, pronouns are often identified as keyphrases (a prime example being the pronoun "я" ("I") in texts narrated in the first person) (Svetozarova and Shtern, 1989).

For scientific texts, the following features are important (Moskvitina, 2009).

- High frequency. The occurrence frequency within the studied text is higher than in the language in general.
- Semantic proximity to the text's theme. Connection to the text's meaning, e.g., synonymy, hyponymic/hypernymic relationships.
- Informational density. Maximum amount of information with minimal volume; consequently, terms often function as keyphrases in scientific texts.

Synthesizing the various interpretations proposed by researchers (e.g., the previously mentioned Svetozarova and Shtern, 1989; Shekhtman, 2005; Moskvitina, 2009), it can be concluded that generally keyphrases inherently possess a set of common features, which typically encompass the following.

- Significance for text comprehension.
- Semantic integrity (cohesion/wholeness).
- Mutual intelligibility across a broad readership (or professional community).
- Conciseness and the capacity to represent substantial volumes of information compactly.

The aforementioned attributes are relevant to both written and spoken textual modalities.

**2.3. Specific Features of Keyphrases in Spoken Text**

Keyphrases in spoken text can possess all the features they exhibit in written text, but with the addition of several peculiarities related to the production and perception of spoken language. Consequently, in spoken text, keyphrases are characterized by "prosodic specificity" ("просодическая специфичность") (Yagunova, 2004). Significant features of keywords in spoken text include phonetic attributes such as their position within a syntagma and the presence of a pause before and after the word.

Research findings (Yagunova, 2004) indicate that keywords are 40% more likely to occur at the end of a syntagma than other text components. The probability of a pause before other words is 70% higher than before keywords. Concurrently, the probability of a pause occurring after a keyword is 1.5 times greater than the probability of it occurring after other words, and almost twice as great for pauses exceeding 750 ms (a duration corresponding to the end of a sentence). An experiment was also conducted with 8 participants whose task was to reconstruct a text after listening with interference (under white noise conditions). Participants recognized keywords 20% better than non-keywords. Furthermore, their recognition accuracy in the final syntagmatic position was 33% higher. Pauses of less than 750 ms, implemented before words, increased keyword recognition accuracy by 23%, and by 55% when such pauses occurred after keywords. Analogous effects for other words were practically absent.

The acoustic (including prosodic) features inherent in keyphrases are not sufficiently studied. A significant number of studies focus on extracting keyphrases from text, but far fewer works are based on spoken texts, and even fewer studies utilize information from audio signals, such as acoustic features. Automatic extraction of keyphrases from spoken language will be discussed in more detail in Section 2.4.

Chen et al. (2012) propose using the following groups of acoustic features:

● Duration-related features. The authors postulate that the duration of words when functioning as keywords may exceed their duration when they are not keywords. Based on this, the average duration of each phonetic unit from a corpus is calculated, and then the duration is normalized by the mean value. Subsequently, for each keyword, its maximum, minimum, average, and range of normalized values are used as duration characteristics.

● Fundamental frequency (F0) features. It is validated that keywords may be distinguished by a wider fundamental frequency range compared to non-keyphrase segments. As with duration, maximum, minimum, average, and range of F0 values are calculated. This measurement is highly relevant because the fundamental frequency is an acoustic correlate of vocal fold vibration and functions as a crucial marker reflecting both the speaker's affective state and the language-specific contours observed in spoken discourse.

● Spectral energy features. It is also confirmed that keywords are characterized by higher energy levels. In this regard, the authors (Chen et al., 2012) rely on cepstral coefficient data. The maximum, minimum, average, and range values are also calculated as characteristics.

Although the authors generally report positive results confirming their hypotheses, it is important to note that the study was conducted on English and Mandarin Chinese. Consequently, results may differ for other languages. Furthermore, acoustic features were examined not in isolation but in conjunction with lexical and semantic characteristics.

Among the phonetically significant features for the Russian language, Yagunova (2004) reliably confirms only the role of pauses as potential indicators of keyphrases in a text, while the connection between keyphrases and other acoustic features remains unclear.

Guseva et al. (2024) investigate the acoustic characteristics of keyphrases manually identified by participants in spoken popular science texts. An analysis using the OpenSMILE library and the eGeMAPS v0.2 feature set reveals the presence of acoustic correlates associated with keyphrases identification. However, only weak positive or negative correlations are found between individual acoustic features and the keyphrases identified by participants. No significant correlations are found with the keyphrases' position in the sentence or pauses before/after them. A weak but consistent correlation is noted with spectral and frequency characteristics, suggesting that Russian keyphrases might be characterized by a wider pitch range and greater energy, which aligns with findings from studies on other languages (Chen et al., 2012).

The features that may be significant when identifying keyphrases in a spoken text are illustrated in Figure 6.

**Figure 6.** Characteristics of keyphrases in spoken text
**Рисунок 6.** Характеристики ключевых выражений в устном тексте

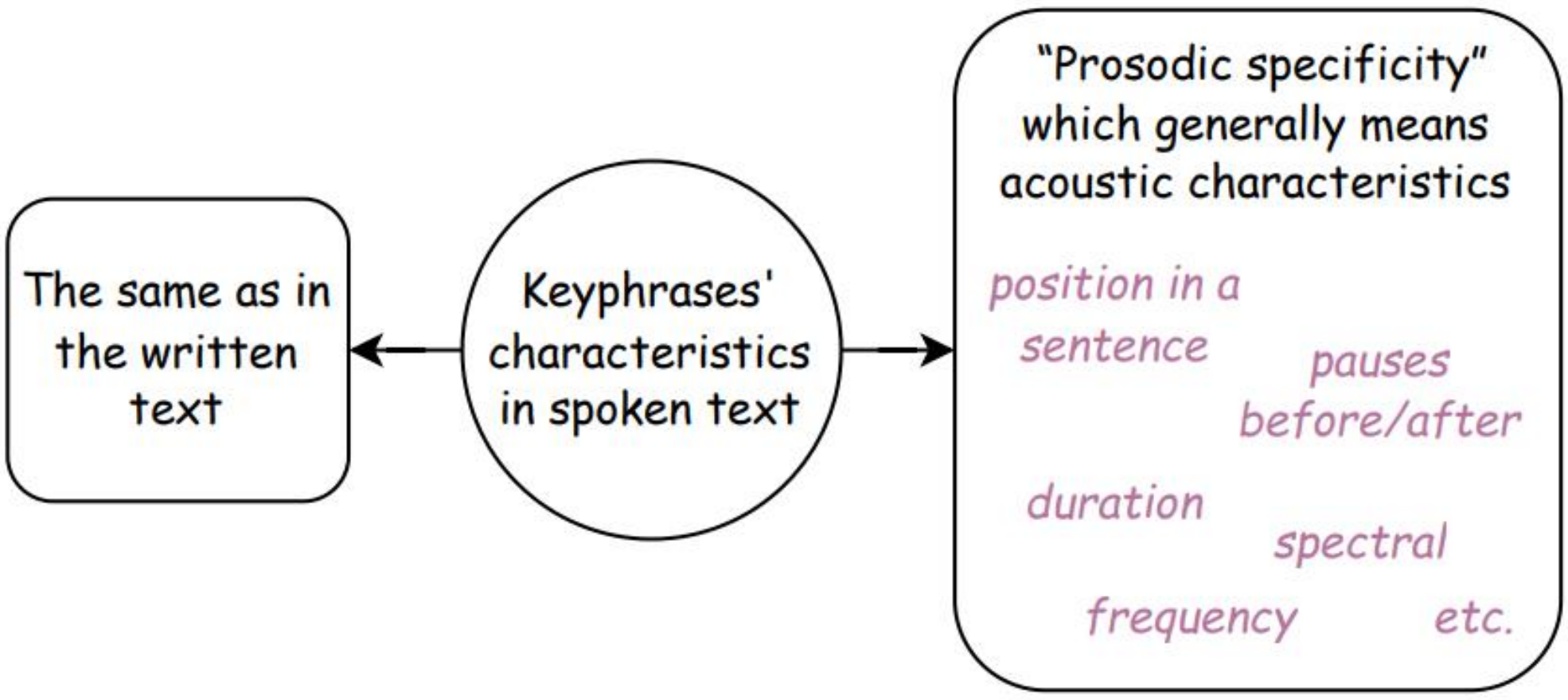

**2.4. Functions of Keyphrases**

Research has described several primary functions of keyphrases, which are listed below in a progression that moves from macro-level text organization and system utility toward individual user memory and comprehension.

● Highlighting essential information. This allows distinguishing a text from others, classifying it, and assigning it to a specific thematic area (Gulyaev and Lukashevich, 2013).

● Information structuring. Information becomes more understandable and easier to perceive (Mitrofanova and Gavrilik, 2022).

● Document clustering. This facilitates grouping documents into relevant categories and simplifies information retrieval (Popova and Danilova, 2014).

● Content evaluation of documents. This information can be applied in algorithms for indexing, categorization, summarization, and document compression (Dubinina, 2020).

● Memory storage. The capacity of human working memory corresponds to the number of keyphrases typically comprising a set. This facilitates memorization and recall of information (Moskvitina, 2018).

Among the functions of keyphrases accompanying scientific articles, the following are particularly highlighted.

● Determining the subject area of a scientific publication.

● Enabling retrieval of a desired article from a database via a search query.

● Gaining an understanding of article similarity without detailed content review (Abramov, 2011).

● Communicative function. Keyphrases help organize subject catalogs and scientific journals, simplifying the search for necessary information and accelerating the process of scientific communication (Moskvitina, 2009).

Thus, keyphrases perform several important functions, which are schematically illustrated in Figure 7.

**Figure 7.** Primary functions of keyphrases
**Рисунок 7.** Основные функции ключевых выражений

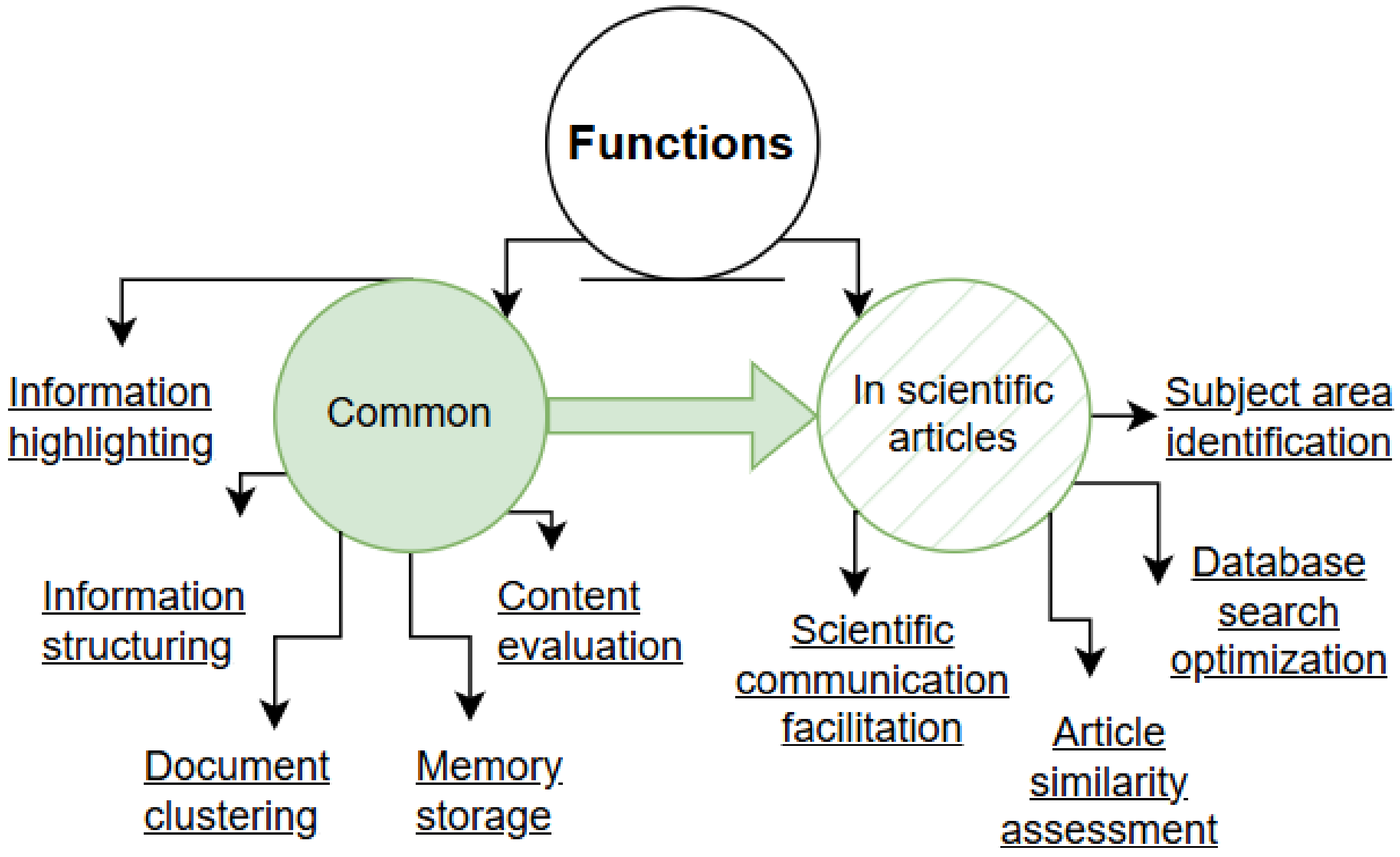

**2.5. Frequency and Distribution of Keyphrases in the Original Text Structure**

The selection of keyphrases is based on their frequency within a specific text, not the language as a whole (Bolshakova et al., 2011). For example, in an experiment using linguistic articles, participants without linguistic education who poorly understood the meaning of the presented texts identified linguistic terms that appeared most frequently in the texts as keyphrases (Sakharny and Shtern, 1988). In literary texts, words denoting the subject of description and/or characterizing characters and main plot events are often noted as key (Grudeva and Churilina, 2019; Yagunova, 2010).

In perceptual experiments, informants tend to include expressions in their sets that are not actually present in the text but convey its main idea (Grudeva and Gubushkina, 2020). Nevertheless, the study by Svetozarova and Shtern (1989) confirms that, despite the high variability in the sets of keywords, a core emerges from the participants' responses, consisting of words chosen by all participants.

The uneven distribution of keyphrases within the structure of the original text also warrants consideration (Sakharny and Shtern, 1988). Words and collocations present in the title, abstract, introductory, and concluding sections of a text possess particular informativeness. This distribution is linked to the fact that these sections contain key information that facilitates rapid comprehension of the text's main idea and content (Kodzasov and Krivnova, 2001; Vasilyeva and Konkov, 2015). This is particularly characteristic of short texts, which are often used in perceptual experiments, where the theme is usually established at the beginning. Cases where keyphrases first appear at the end or in the middle of the text are less frequent. However, if substitute words (synonyms, pronouns, etc.) and the fact that words can appear repeatedly in the text are taken into account, their distribution becomes more even (Svetozarova and Shtern, 1989).

**2.6. Keyphrase set size**

Many researchers, when deciding on the necessary volume of keyphrase sets for a text, rely on the approach proposed in (Sakharny and Shtern, 1988). Conducted experiments show that participants can work with virtually any number of expressions, but the most convenient predefined range, especially when dealing with spoken text, is a set of 7-10 words or collocations. Based on this, researchers recommend extracting 5 to 15 expressions depending on the text volume (Sakharny and Shtern, 1988). An extension of this approach is the identification of small, medium (sometimes), and large sets of keyphrases (Svetozarova and Shtern, 1989). Results obtained from analyzing 20 texts of varying sizes (40-400 words), structures, and genre-stylistic affiliations indicate that the sets identified by auditors range from 5 to 20 expressions.

Various publishers and conference organizing committees typically recommend authors to provide 3-5 to 15 terms (Kamshilova, 2013). For example, when submitting an article to the international scientific journal Terra Linguistica[1] authors are required to provide 5 to 7 keyphrases.

While each additional keyphrase offers an opportunity to attract readers to an article (Abramov, 2011), accuracy in their selection is equally important. The chosen elements must carry fundamental information about the document (Kamshilova, 2013).

Thus, the boundaries for keyphrase sets are generally established within the range of 3 to 15 expressions (see Figure 8).

**Figure 8.** Keyphrase set size
**Рисунок 8.** Объём набора ключевых выражений

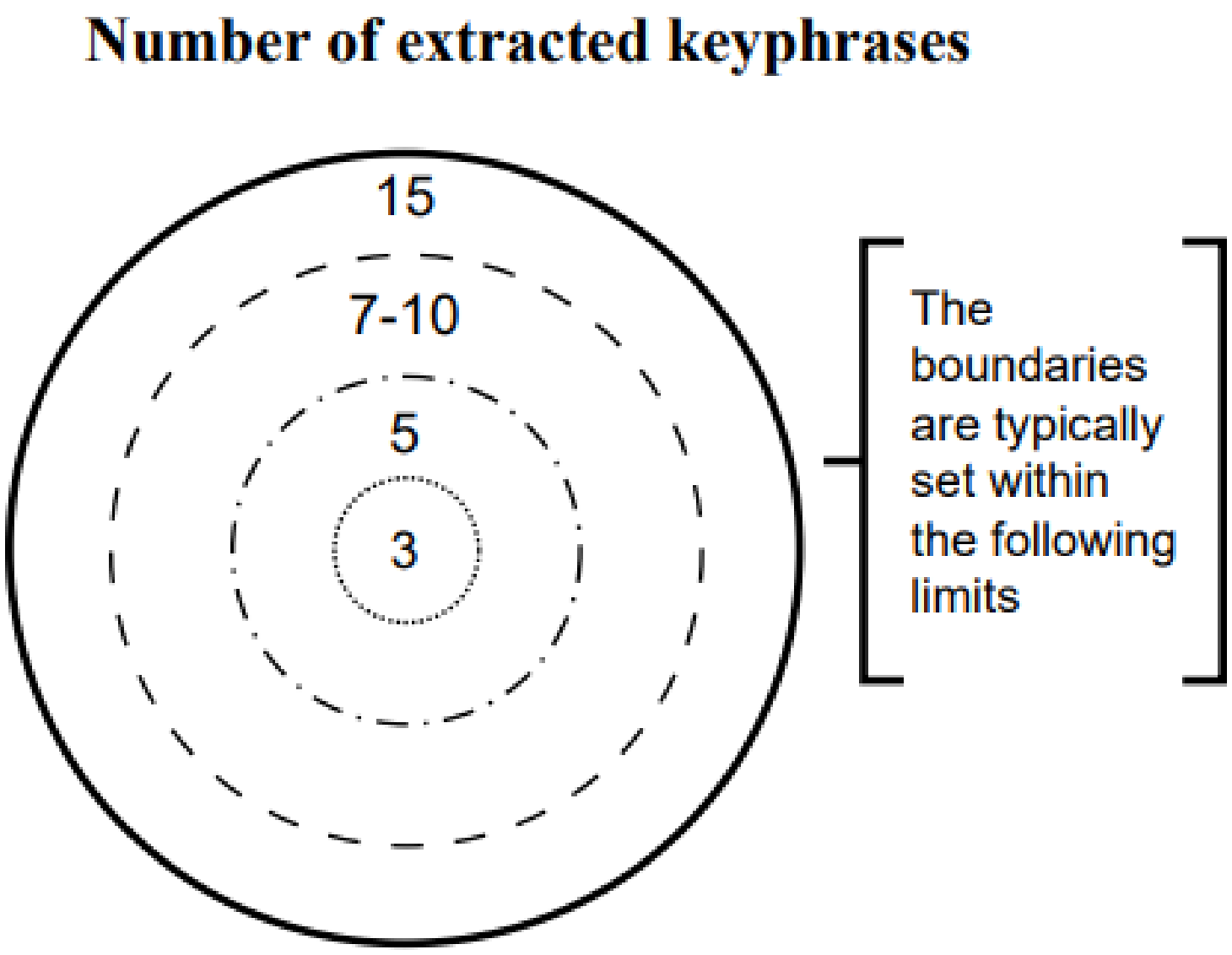

[1] Available at: https://human.spbstu.ru (Accessed 22.11.2025)

### 2.7. Manual and Automatic Keyphrase Extraction

Keyphrases can be extracted manually or automatically. In the former case, extraction can be performed directly by the text's authors (e.g., when submitting an application for a scientific conference), by subject matter experts, or by participants (Vanyushkin and Grashchenko, 2018). Schematically, the methodology for manual keyphrase extraction involves the following: after familiarizing themselves with a document, a participant selects a few words that, in their opinion, convey the main content of what has been read (or heard) (Sakharny and Shtern, 1988).

Discrepancies can arise where keyphrases identified by participants or even authors are either sparsely represented or entirely omitted within the source text (Mitrofanova and Gavrilik, 2022). In such cases, we speak of "present" and "absent" keyphrases in the text (Chen et al., 2020).

Manual keyphrase extraction is a highly labor-intensive process. Consequently, it is poorly suited for processing large volumes of text and working with corpus data. Furthermore, human selection is always subjective, as it depends on the reader's/listener's interpretation of the text.

Automatic keyphrase extraction, in a simplified sense, involves selecting words and collocations that most accurately convey the content of the document under consideration. In essence, this refers to the automatic identification of non-randomly occurring elements that are significant for the sample in the context of the overall dataset (Bolshakova et al., 2011). The extraction process occurs without human intervention and depends on the chosen model (Zhang et al., 2008). The significance level of an expression as a keyphrase is determined by its relative frequency of occurrence in experimental participants' responses (for manual extraction) or is calculated based on a specific algorithm (for automatic extraction) (Bolshakova et al., 2011).

Automatic keyphrase extraction methods are analyzed in detail in Section 3.

## 3. Automatic Keyphrase Extraction

### 3.1. Overview and Limitations of Automatic Keyphrase Extraction in Written Text

Automatic keyphrase extraction from text plays a significant role in Natural Language Processing (NLP). It is applied in tasks such as creating automatic information retrieval systems, document classification and categorization, content summarization and annotation, and others.

In recent decades, the volume of research in the field of automatic KE has significantly increased due to the greater availability of computational and informational resources (Vanyushkin and Grashchenko, 2017). Various automatic methods for keyphrase extraction exist, including statistical, hybrid (linguostatistical), machine learning-based, and structural approaches (classifications will be discussed in Section 3.3.). However, despite the diversity of approaches, the process of automatic keyphrase extraction still faces several challenges.

Hasan and Ng (2014) note the following factors influencing the complexity of keyphrase extraction.

- Volume of the source text. The longer the text, the more potential keyphrases there are, and the more difficult it is to extract them.
- Structural consistency. While keyphrases typically anchor to specific sections within structured documents, the disruption of this expected structural integrity challenges extractive algorithms dependent on positional cues.
- Topic shift within the document. If the subject of discussion changes, the keyphrases associated with the topic also change.
- Interconnectedness. Keyphrases in scientific and news articles are usually interconnected, but in informal texts, topics within a single document may be unrelated, complicating extraction.

The automatic extraction of multi-component keyphrases, particularly of certain types of lexical categories, presents a particular challenge (Sheremetyeva and Osminin, 2015). For example, noun phrases can include various parts of speech (nouns, adjectives, numerals, or their combinations). Consequently, correctly extracting keyphrases requires consideration of the morphosyntactic and semantic characteristics of the language.

Furthermore, automatic methods often require a large amount of training data. For the correct operation of ML algorithms, thematic text corpora with annotated keyphrases are necessary. Although the use of corpora enhances the accuracy of the keyphrase extraction procedure, applying such methods is often complicated by the absence of a corpus for a specific subject area or language.

Statistical algorithms also encounter different kinds of limitations in the task of automatic keyword extraction. While they do not require specialized linguistic knowledge bases and operate relatively quickly, they often yield unsatisfactory results. The scope of effective use for statistical methods is limited to morphologically poor languages, where the frequency of different word forms for a single lexeme is relatively high due to a comparatively small number of grammatical changes and inflections. These languages include, for example, English, Danish, Modern Chinese, and Vietnamese (Tsarfaty et al., 2013). Due to this characteristic, statistical methods may not be suitable for working with languages that have rich morphological systems (Russian, French, German, Arabic, Turkish, and others), where each lexeme has numerous word forms with low frequency in any given text.

Methods for automatic keyphrase extraction are developed for specific tasks, application domains, and languages. For instance, many available algorithms are designed for the English language and therefore require adaptation when working with other languages. Moreover, the choice of method depends on the functional style of the text. For example, literary texts may contain a large number of metaphors, which are difficult to extract using statistical methods. This suggests that there is likely no universal method that can effectively work in all situations.

Thus, the applicability of various automatic keyword extraction algorithms is limited by a number of factors. To create effective methods, it is necessary to consider the typological characteristics of natural languages, subject domains, as well as the availability of linguistic and software resources.

### 3.2. Stages of Automatic Keyphrase Extraction in Written Text

The study by Vanyushkin and Grashchenko (2016) outlines three sequential stages for keyphrase extraction algorithms.

The first stage is text preprocessing. Preprocessing should be performed at all levels (graphemic, morphemic, lexical). Its goal is to represent the text in a format conducive to subsequent keyphrase extraction (Troshina, 2025). Differences arising at this stage are related to the necessity of working with different languages and/or are dictated by the specifics of a particular task. The following procedures may be applied to the text: tokenization (identifying word forms in the text), lemmatization (reducing various word forms to a single representation – the base form or lemma), markup removal, lexical normalization, part-of-speech tagging, and the use of a stop word list (removing a predefined list of stop words that lack semantic load, either generally or for a specific text). Performing text preprocessing procedures primarily involves the use of linguistic databases and dictionaries, which complicates the automation of this stage.

The second stage involves the recognition (classification) of keyphrase candidates obtained in the previous stage, or the filtering of candidates. The recognition procedure, as well as the use of various corpora and dictionaries, depends on the chosen automatic algorithm. Consequently, the procedures performed at the second stage may or may not be language-dependent.

The third stage consists of the postprocessing of the obtained set of keyphrases. At this stage, the set is brought into the required format – truncated, ranked, ordered, etc.

A similar universal scheme for text keyphrase extraction is proposed in the work by Vinogradova and Ivanova (2016). The first stage involves text preprocessing. The second and third stages involve the selection of keyphrase candidates and their filtering according to predefined criteria (which depend on the chosen automatic method). The fourth stage is the final selection of keyphrases from the set obtained after completing the previous steps.

In the study by Sheremetyeva and Osminin (2015), the procedures of forming and subsequently filtering a list of potential keyphrases are identified as common to various automatic algorithms. The extraction and ranking of keyphrase candidates are also discussed in the research by Wienecke (2020).

Thus, in general terms, the principle of keyphrase extraction using various automatic methods follows a unified logic, involving the sequential execution of text preprocessing procedures, keyphrase classification, and their presentation in a format determined by the task (see Figure 9).

**Figure 9.** Stages of automatic keyphrase extraction in written text
**Рисунок 9.** Этапы автоматического извлечения ключевых выражений из письменного текста

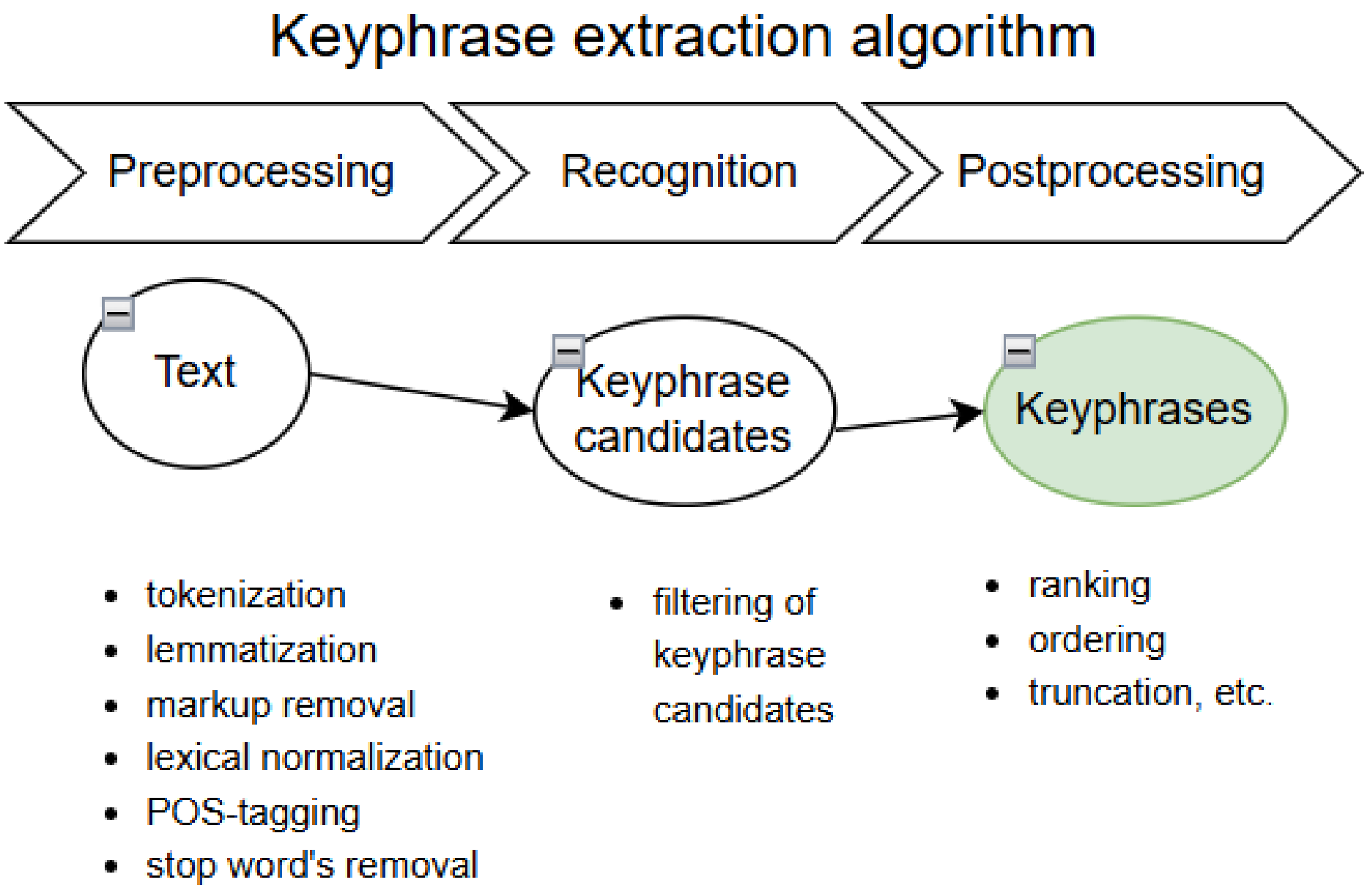

### 3.3. Classifications of Automatic Keyphrase Extraction Methods in Written Text

This section reviews several classifications of automatic keyphrase extraction methods in written text. A number of researchers divide automatic algorithms into just two main groups, including statistical and machine learning-based (Chen and Lin, 2010), or statistical and hybrid, with the latter further considering sub-groups of graph-based and ML-based methods (Sheremetyeva and Osminin, 2015).

Vanyushkin and Grashchenko (2016) propose a more granular classification based on keyphrase recognition systems. Such systems categorize input data into two classes based on whether an element belongs to the keyphrases. Based on this, they suggest classifying methods along three facets by:

● type of learning (unsupervised, self-learning, and supervised)

● type of linguistic resources required (without usage, dictionary-based, ontology-based, based on annotated and unannotated corpora)

● type of mathematical apparatus of the keyphrase recognition system (statistical, neural network-based, hybrid, structural; the latter are further divided into graph-based and syntactic).

Morozov et al. (2023) also discuss classification by type of learning. Keyphrase extraction algorithms are divided into unsupervised and supervised learning methods. The first group of methods uses features to rank words in a text and select the required number of words with the highest rank. Statistical and graph-based methods operate in this manner. Supervised learning methods require a training set consisting of texts and a gold-standard list of keyphrases extracted from them (the set of standards is usually compiled by an expert) for preliminary algorithm tuning.

The authors of a previous study (Koloski et al., 2021), followed a similar approach, using supervised machine learning techniques to extract keyphrases from materials not only in the Russian language, but also in several low-resource languages (Estonian, Croatian, and Latvian).

In the subsequent analysis, we adhere to the classification of automatic keyphrase extraction methods into statistical, hybrid (linguostatistical), machine learning-based, and structural (graph-based). We then examine each of these groups in more detail and provide examples of specific algorithms.

It is important to acknowledge that the categorization of KE methods into distinct groups is inherently somewhat arbitrary, as many modern approaches combine elements from multiple paradigms. The categories we employ should therefore be understood as useful heuristics highlighting dominant characteristics rather than mutually exclusive boxes.

#### 3.3.1. Statistical Methods for Keyword Extraction

As previously indicated in Section 3.1, statistical extraction methods are subject to inherent limitations. These methods are based on analyzing the frequency of words and their combinations within a text. While this approach does not necessitate prior training and allows for facile application to novel data, rendering them versatile, frequency alone is not always the most decisive factor in identifying the most important expressions for a text, which means these methods may be less accurate than those considering semantics and context (Sheremetyeva and Osminin, 2015). Nevertheless, the relevance of statistical methods increases when studying new subject areas and their terminology (Vinogradova and Ivanova, 2016).

In its simplest form, the overall frequency of word forms, reduced to their root (lemma) through stemming, is calculated. Word forms that are ranked by frequency and exceed a given threshold are considered to be keywords (Sheremetyeva and Osminin, 2015).

Chronologically, the first methods for automatic keyword extraction, discussed for instance in works by Piotrovsky et al. (1977) and Luhn (1957), were based on ranking all word forms (lexemes) found in a text by their frequency and selecting keyword candidates based on this principle.

Statistical methods include TF-IDF, Chi-squared, Log-Likelihood, PMI-test, T-test, and others.

The TF-IDF (Term Frequency-Inverse Document Frequency) measure is one of the most common statistical methods (Jones, 1972). The TF-IDF weight depends on how frequently a given expression appears in a specific document relative to its frequency in other documents within a corpus. The higher the frequency in the document under investigation and the lower it is in others, the greater the TF-IDF weight. Consequently, this approach allows for assessing the significance of a word or phrase for understanding a document's content within the analyzed corpus.

The Chi-squared criterion quantifies the statistical significance of frequency differences. Unlike TF-IDF, it can be applied to a single text or a small sample of texts without requiring additional corpora or a list of comparison documents. The method's principle involves comparing observed frequencies with expected frequencies based on a hypothesis. A significant difference between observed and hypothesized frequencies indicates that the hypothesis is not supported. Constructing a co-occurrence matrix of expressions within a text enables a more precise estimation of the statistical significance of frequency differences. It also reduces the influence of low-frequency words that might receive disproportionately high scores due to matrix sparsity (Krasavina and Mirzitova, 2015).

The Log-likelihood association measure is typically used to quantify the degree of systematic association between parts of collocations (Zakharov and Khokhlova, 2014; Brezina et al., 2015). Based on the actual parameter values of data and their probabilistic model, this method yields parameter values that most accurately correspond to reality (Mitrofanova and Gavrilik, 2022). When applied to keyword extraction, the Log-likelihood association measure assesses how often words actually co-occur in a text compared to their random co-occurrence probability. If the probability of joint occurrence is significantly higher than expected by chance, it may indicate that these words form a keyword expression.

The PMI-test (Pointwise Mutual Information) metric measures the degree of association between an expression and a text or corpus. A high PMI score suggests that the expression frequently appears in the document or corpus and may therefore be a keyword.

The T-test method determines whether there is a significant difference between the average frequency of an expression's occurrence in the studied text and a control text. A significant difference indicates that the expression is a keyword for the text.

**3.3.2 Hybrid (Linguostatistical) Methods for Keyword Extraction**

Hybrid methods for keyword extraction combine statistical and linguistic approaches, leveraging the advantages of both. Such algorithms can produce more accurate results than purely statistical methods because they consider various features for assigning keyword status to expressions:

- semantic (e.g., analysis of semantic relationships between words, such as synonymy, antonymy, and hyponymy);
- syntactic (e.g., analysis of sentence syntactic structure);
- morphological (e.g., part-of-speech tagging);
- lexical (e.g., frequency of expression usage in a specific domain or text type).

Furthermore, some hybrid methods can incorporate features specific to particular domains or text types. For instance, when extracting keywords from medical texts, the presence of medical terminology can be considered. However, hybrid methods are more complex and computationally intensive. Their effectiveness can also strongly depend on the quality of the training data used.

The RAKE (Rapid Automatic Keyword Extraction) method is domain-independent and does not require prior model training or corpus usage. It assigns weights to each word in a text based on its frequency and position. The original RAKE algorithm was developed for English (Rose et al., 2009). Versions adapted for Russian exist, accounting for features, such as cases and declensions (Moskvina et al., 2017). The algorithm performs syntactic analysis and tokenizes the text by punctuation and stop words. In the Russian-adapted algorithm, grammatical rules of a Russian-language syntactic parser based on NLTK (NLTK4RUSSIAN) can be used to define syntactic group boundaries. For each token, the following metrics are calculated: its frequency in the text and its degree of co-occurrence with other tokens (co-occurrence measure). After calculating these metrics, each token is assigned a weight, computed as the ratio of its occurrence frequency to its co-occurrence degree. Tokens with the highest weights are considered semantically more significant for the document and are identified as keyphrases.

The YAKE (Yet Another Keyword Extractor) algorithm (Campos et al., 2020) is based on RAKE principles. However, it does not use stop words for tokenization and considers a greater number of metrics for calculating keyword weights:

- normalized word frequency;
- word position in the text;
- number of sentences containing the token;
- number of capitalized occurrences (casing);
- word relatedness to context; it is assumed that if a token has a high degree of similarity, it is less important for the context.

Keyphrases are selected based on weight ranking. Since YAKE extracts ready-made and frequent fragments from a text, it is suggested for use in summarization and similar tasks (Morozov et al., 2023).

For English texts, the Topia algorithm is used, extracting keywords based on text tokenization and morphological analysis of a corpus. For Russian, the analogous library RuTermExtract has been developed. The primary difficulties encountered when working with it are incomplete rules and ambiguity during morphological parsing.

Hybrid methods also include SpaCy, an advanced library for Natural Language Processing. This library enables keyphrase extraction by considering word frequency or the weight of distributed vector embeddings. Additionally, there is the linguistic toolkit PullEnti, which extracts keywords based on rules that consider morphological (morphological parsing is performed for each token separately) and semantic features.

Overall, the application of hybrid KE methods, combining statistical and linguistic approaches, allows for accounting for a greater number of factors and improving the accuracy and completeness of results compared to using only one of these methods.

**3.3.3. Machine Learning-Based Methods for Keyphrase Extraction**

Machine learning-based algorithms are sometimes categorized as a subgroup of hybrid methods (Sheremetyeva and Osminin, 2015). Such methods typically require a corpus with pre-labeled keyphrases to create a training set and build a classifier. During training, keyphrases are marked as positive examples, and all other elements are marked as negative. Word vectors are generated based on specific parameters (e.g., TF-IDF measure, word length, word position in the title/first paragraph/last paragraph). The probability of a word combination belonging to the set of keyphrases is calculated, and a threshold for inclusion is set. After training, the model computes word relevance based on parameter vectors and determines the probability that a word combination expression is also a keyphrase.

ML-based methods include decision trees (Moskvina et al., 2018), Bayesian methods (Rose et al., 2009; Sheremetyeva and Osminin, 2015), Support Vector Machines (Vanyushkin and Grashchenko, 2016), and neural networks (Abrosimov and Mosyagina, 2022).

KeyBERT is a contextualized predictive model of distributed vectors, based on the original BERT model (Devlin et al., 2019). The method relies on a bidirectional transformer that converts sentences and texts into vectors to reflect their meaning. KeyBERT determines the cosine similarity of potential keyphrase vectors relative to the text in which they are presented and uses pre-trained models to calculate word relevance in context (Grootendorst, 2020). First, the text is processed using a pre-trained BERT model. Each word in the text is transformed into a feature vector representing its semantic meaning. Next, the algorithm calculates word relevance by comparing their feature vectors with the feature vectors of other words in the text. Relevant words are grouped into clusters, each representing a potential

keyword expression. From each cluster, the most representative word is selected and combined with other relevant words to form a keyphrase.

In addition to KeyBERT, other ML-based methods for keyphrase extraction exist. The KEA algorithm is quite common, based on a probabilistic classification model (Moskvina et al., 2017; Moskvina et al., 2018). Its effectiveness for the Russian language has been clearly demonstrated in a study by Sokolova and Mitrofanova (2018).

### 3.3.4. Structural (graph-based) Methods for Keyphrase Extraction

Structural methods, sometimes considered within hybrid KE approaches (Sheremetyeva and Osminin, 2015), are based on analyzing the text's structure and the relationships between its elements. These methods employ various algorithms to construct word graphs based on their relative positions in the text and identify keyphrases based on their significance within the graph. The vertices of the graph represent keyphrase candidates, and the edges connecting them are weighted according to the proximity of these candidates.

Methods differ in their candidate selection and proximity determination techniques, which are based on statistical, morphological, syntactic, and sometimes semantic analysis. One common structural method is TextRank. It is based on the PageRank algorithm, used by Google's search engine for ranking web pages (Mihalcea, 2004). The core idea of TextRank is to form a weighted graph where nodes represent tokens, lemmas, or phrases, and edges reflect textual relationships, with weights indicating the strength and/or type of semantic connections. Keyphrase candidates with high ranks are considered more significant.

Other graph-based approaches include, for instance, the language-independent algorithm DegExt. In DegExt, stop words are filtered before graph construction. In the graph, edges between vertices connect only words that are adjacent in any sentence and are not separated by punctuation (Litvak, 2013).

Some structural algorithms can utilize hypergraph models. In a hypergraph, any subset of vertices can form an edge. Such methods can be effective for large texts but are complex to implement and require significant computational resources.

### 3.3.5. Keyphrase generation

Keyphrase generation is a new approach that has emerged with the advent of Large Language Models (LLMs). LLMs can enrich and expand the keyphrase sets assigned to the text and predict topically determined lexical units that are observed or unseen in a given text. The applicability of keyphrase generation was evaluated primarily for the English datasets (Boudin, Aizawa 2025). Figure 10 shows that pretrained LLMs efficiently predict keyphrases present or absent in the texts, the core difference from keyphrase extraction being operations on embedding vectors.

**Figure 10.** Keyphrase prediction process (Umair et al., 2024)
**Рисунок 10.** Процесс предсказания ключевых выражений (Umair et al., 2024)

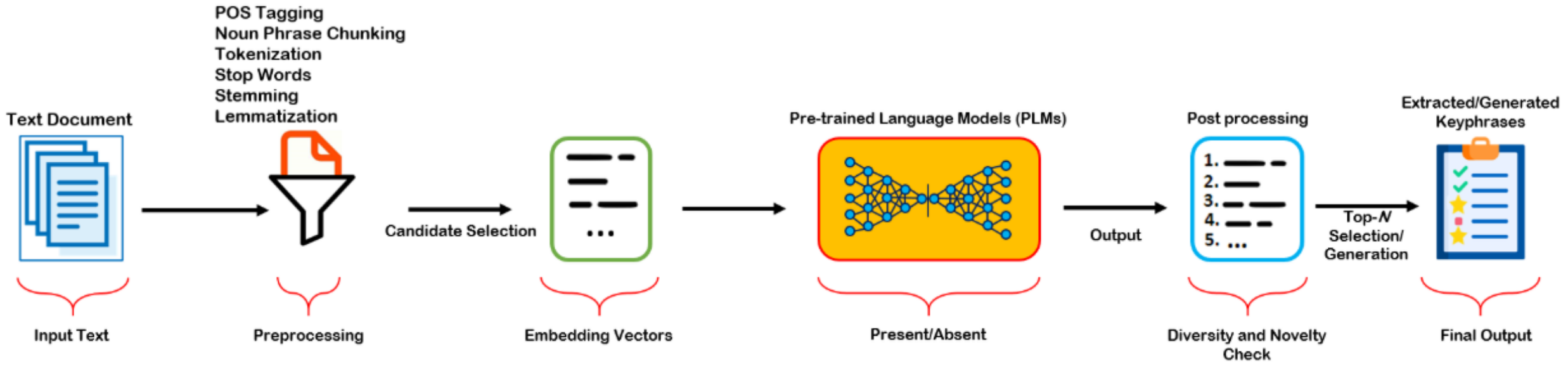

In recent works promising results were obtained by zero-shot and few-shot prompts for keyphrase generation by models from the GPT family. The paper (Martínez-Cruz et al., 2023) focuses on GPT models advantages in keyphrase generation taking into account domain adaptation and long document processing. The development of prompt engineering methods has shown advantages of zero-shot and few-shot prompts in keyphrase generation (Song et al., 2023). It was shown that the choice of prompt type and design improves the results, with a few-shot approach being more effective than zero-shot prompting.

Several studies give insights into hybrid techniques combining a set of instruction-based models or adjustment of instruction-based models and other approaches. In (Wang et al. 2024) a complex procedure for keyphrase generation is introduced, combining a zero-shot learning approach for knowledge graphs with LLMs, which includes keyphrase extraction, extension, retrieval and ranking. In (Shao et al., 2024) the authors introduced a one2set-based model as a keyphrase generator producing candidates which are further ranked by LLM select keyphrases from these candidates.

In Russian NLP prominent results were obtained by the NeuroRNC group (Glazkova et al., 2024; Glazkova, Morozov, 2024; Glazkova et al., 2025). The researchers introduced novel approaches towards keyphrase annotation of documents in large corpora and evaluated performance of classical techniques (TF-IDF, RAKE, KEA, TextRank, etc.) in the task of keyphrase extraction as well as LLMs for keyphrase generation in news texts and research papers. Recent advances deal with temperature effects and robustness of LLMs in keyphrase generation.

## 4. Automatic Keyword Annotation

The expansion of the information space and the necessity of analyzing extensive corpus data necessitate the automation of text indexing. Annotating keyphrases in texts placed in electronic collections is crucial for users to discover them via search engines. The implementation of the Semantic Web project, which practically applies semantic markup standards using formal ontologies, along with Knowledge Graph technology, elevates the indexing task to a new level by specifying keyphrases as domain concepts, named entities, facts, relationships, and so on.

Modern corpus research involves working with diverse types of corpus annotation. Alongside fundamental annotation types (morphological, syntactic, pragmatic, word-formation), there are also applied types (discourse annotation, anaphora annotation, sentiment annotation, entity annotation, genre-thematic annotation, etc.).

Annotating keyphrases in text corpora allows for a content-driven approach to corpus searching, similar to web document searching. The goal of simple corpus search is to find all contexts with a target word or phrase. However, searching by keyphrases allows for forming a selection of documents where the query word is not accidental, peripheral, or insignificant to the text’s content, but rather part of the text’s semantic core. This is possible precisely because sets of keyphrases can convey text content in a concise and structured manner.

Corpus resources with keyphrase annotations are of high value, yet they are currently scarce.

In the study (Hulth, 2003) automatic keyphrase extraction is treated as a supervised learning task. The dataset used in the experiments consists of 2000 abstracts for journal articles on information technology in English with corresponding titles and keywords from the Inspec database. Each abstract has two sets of manually assigned keywords: a set of controlled terms, i.e., terms restricted to the Inspec thesaurus, and a set of uncontrolled terms. A series of classification models was trained using n-grams, part-of-speech patterns, noun phrase boundaries, statistical features. It was shown that the best result in precision was shown by the model trained with noun phrase boundaries, in recall – by part-of-speech patterns, and in the F-measure – by n-gram models.

The study (Nhuen and Kan, 2007) presents NUC corpus including texts from 120 scientific articles with annotated keyphrases. The authors used KEA as a baseline which was improved in a modified algorithm that considers keyphrase position in the text, part-of-speech tagging, and the use of acronyms and affixes typical of scientific terminology. The models were trained on the corpus with pre-annotated keyphrases. An unconventional technique was used in classification: a vector was formed with coordinates corresponding to the keyword frequency in different text sections (title, abstract, introduction, main sections, conclusion, references, etc.). That allowed it to significantly exceed the baseline.

The project by Wan and Xiao (2008) deals with the DUC2001 corpus used as a baseline for summarization. The corpus includes 308 news articles representing 30 topics. The average text length is around 740 tokens. Manual annotation was performed by 2 experts, with about 8 keyphrases per text, and an average keyphrase length of about 2 tokens. Automatic annotation used a hybrid approach: clusters of topically similar documents were formed taking into account cosine similarity of averaged text vectors, keyphrases were extracted within the clusters by means of ExpandRank algorithm, and then a set of keyphrases for individual documents within a cluster was identified by SingleRank algorithm. Keyphrase weights were determined as regards corpus-based metrics (frequency, types of lexico-grammatical patterns, etc.).

The study (Krapivin et al., 2010) describes a dataset of 2,000 publications from the ACM Digital Library in the field of information technology. The texts were preprocessed, annotated with OpenNLP, and segmented into text sections. In the experiments, the following models were trained: Support Vector Machine (SVM), Local SVM (FaLK-SVM), and Random Forest. The evaluation shows encouraging results which surpass KEA based on a Naive Bayes Classifier, improving the average F-measure from 22% (KEA) to 30% (Random Forest) on the same dataset without using controlled vocabularies. The study demonstrates the importance of affixes and acronyms as keyphrase markers.

Mijić et al. (2010) present a keyphrase extraction algorithm implemented in a search engine for Croatian news texts. The algorithm is characterized by its simplicity, speed, and efficiency, involving candidate keyphrase identification considering word form and construction frequencies, lemmatization, and candidate ranking using the TF-IDF metric. To evaluate the algorithm's quality, a corpus with manual annotation was developed, including 1020 news texts annotated by 8 annotators. Keyphrase sets generated automatically and manually were compared across 4 parameters: exact keyphrase match, partial match (morphological variants, inclusion, etc.). The results indicate that automatic keyphrase annotation improves with the ability to filter keyphrase candidates by part of speech and lexico-grammatical patterns.

The research by Marujo et al. (2013) proved the significance of semantic and rhetorical features of keyphrases in a text corpus for their automatic annotation. For experiments, texts from 10 topics (technology, crime, sports, health, culture, fashion, science, business, world politics, US politics), divided into 85 subclasses, were combined into a corpus. In addition to standard keyphrase features (TF-IDF, text localization, number of word forms in a keyphrase, part of speech), semantic features (number of named entities, number of capitalized word forms, lexico-grammatical patterns, keyphrase frequency in a reference dataset, etc.) and rhetorical features (discourse markers for text continuation, logical inference, contrast, cause-and-effect relations, comparison, emphatic constructions, locations, etc.) were considered. After manual keyphrase annotation via crowdsourcing, 500 texts with the highest annotator agreement (around 90%) were selected. This set was used to train and test a decision tree classifier. The results suggest that the quality of automatic keyphrase annotation is determined by both the topic and the semantic-syntactic and structural characteristics of the texts.

The study (Meng et al., 2017) presents the English-language KP20k dataset, aimed at a more complex formulation of the keyphrase extraction and generation task, as it compensates for the shortcomings of previous approaches and allows the inclusion of not only the observed words of the text in keyphrases but also their semantic correlates (synonyms, thematically related words). The dataset includes 20,000 computer science article texts with sets of keyphrases, titles, and abstracts. This corpus was formed by randomly selecting texts from a corpus of 567,830 articles. The solution for automatic keyphrase generation is implemented based on an RNN model in two versions – a classic RNN encoder and an encoder with a CopyRNN mechanism. A total of 2,780,316 ⟨text, keyphrase⟩ pairs were used to train the model. Testing was conducted on data from the INSPEC, Krapivin, NUS, SemEval 2010 corpora, and the test segment of KP20k. TF-IDF, TextRank, SingleRank, ExpandRank, KEA, and Maui were chosen as baseline methods. In tasks involving both predicting present keyphrases and absent ones, CopyRNN showed the best performance. The CopyRNN dataset and model were also applied to generate keyphrases in news articles of different topics, with results confirming the feasibility of cross-domain transfer.

In 2017, the SemEval competition was held, which included a track on automatic keyphrase extraction (Augenstein et al., 2017). The competition involved evaluating both the quality of the extracted keyphrases and their types and relationships in the original documents — articles on scientific topics. In particular, types of keyphrases such as PROCESS, TASK, MATERIAL, and relationships of synonymy and hypernym-hyponym were considered. The dataset used was a corpus including 100 noun phrases generated from titles and abstracts of articles in physics, computer technologies, chemistry, and computer science. Annotation was performed by 3 annotators using 14 fine-grained types, including PROCESS. The dataset includes 500 texts.

Sterckx et al. (2018) describe a project for annotating keyphrases in a Dutch-language text corpus collected from Belgian media. The corpus comprises four topics, each containing 1,000 to 2,000 texts, including sports news, political news, societal news, and lifestyle news. Keyphrases annotation involved 357 volunteer annotators from various age and social groups. Each text was annotated by multiple annotators, who exhibited low agreement. This did not hinder experiments comparing manual and automatic keyphrase annotation, as it provided a sufficient amount of reference data. Unsupervised machine learning-based methods (TF-IDF, TextRank, Topical PageRank) were used for automatic keyphrase annotation, trained on the corpus considering several parameters: statistical (keyphrase frequency characteristics), structural (text localization), content-related (internal coherence of keywords), and others (lexico-grammatical patterns, inclusion of named entities, additional features from Wikipedia, WordNet thesaurus, etc.). Annotation quality was evaluated by comparing it to reference manual annotations, measured by precision, recall, and F1-score. Experimental results showed that the quality of keyphrase annotation depends on the text’s topic. The best results were obtained for sports and political news segments, while texts from societal and lifestyle news segments showed lower scores for automatic keyphrase annotation quality.

The work (Gallina et al. 2019) describes the KPTimes dataset, which includes about 280 thousand news messages and 10 thousand JPTimes articles with key phrases, annotated by editors. The experiments applied the MultipartiteRank and CopyRNN algorithms (CopySci and CopyNews).

In 2019, the OpenKP dataset was published, including about 100 thousand web documents with expert key phrase annotations (Xiong et al., 2019). The authors developed a new approach to key phrase extraction, namely, BLING-KPE, Beyond Language UnderstandINGKeyPhraseExtraction — automatic key phrase extraction regardless of the domain, based on hybrid word embeddings and a convolutional transformer. The results surpass previous ones.

Russian keyphrase data is involved in few research projects.

Vanyushkin et al. (2019) present the results of keyphrase annotation of a corpus comprising 27 Russian texts from scientific, journalistic, and fiction scientific styles. The annotation was conducted by 24 native Russian speakers. Evaluation of text complexity, perplexity, compression ratio, dictionary richness, and entropy allows to determine lexical richness. As regards keyphrase parameters, researchers concluded that the optimal number of keyphrases is between 5 and 15 units, phrases should not exceed 5 words, and duplicate phrases, isolated function words, pronouns, and adverbs should be excluded from keyphrase sets.

A toolkit for keyphrase extraction from research abstracts was implemented in (Morozov et al. 2023), keyness features were extracted from Kiberleninka electronic library. While working with Russian scientific texts from CS&Math Russian text corpus in the task of keyphrase annotation and generation, the researchers tested a set of algorithms including TF-IDF, TopicRank, YAKE!, RuTermExtract, and a set of Transformer models, namely, KeyBERT, KeyBART, mT5 (Glazkova and Morozov, 2024). Results obtained with mT5 models were taken into account in the development of new RNC functionality for keyphrase search, considering genre-thematic annotation of news texts (Glazkova et al., 2024).

## 5. Review of Current Directions in Automatic Keyphrase Extraction from Spoken Language

With the growth in the volume of multimedia data (videos, music, podcasts, etc.), there is a need for effective approaches to their indexing and browsing. Since multimedia data can contain audio information, automatic keyphrase extraction from spoken language has become relevant (Yun-Nung et al., 2012).

Automatic Speech Summarization (SSum) has wide applications, including extracting key information from news (Maskey and Hirschberg, 2005), podcasts (Vartakavi and Garg, 2020), clinical conversations (Su et al., 2022; Le-Duc et al., 2024), and business meetings (Riedhammer, 2010). Typically, the task of extracting keyphrases from spoken text arises as one of the stages in SSum algorithms.

When raw speech is used as input, the summarization task involves creating a final summary presented in speech or text. Such a summary reduces information volume while providing an understanding of the main content without losing important information (Wang, 2022). For example, in online education, a major difficulty in mastering lecture courses is the lack of time to complete the full course. Utilizing keyphrases extracted from lectures is an effective way for listeners to familiarize themselves with the main concepts (Lee et al., 2014).

Approaches to speech summarization are generally divided into:

- extractive, which select the most important fragments of the source document without altering their structure or content;
- abstractive, which create new text that conveys the main content of the source document in a condensed form, but may differ from the original in structure and style (Wang, 2022).

The relationship between keyphrase extraction and speech summarization is fundamental. KE can serve as both a crucial component and often a primary objective of summarization systems. In extractive summarization, keyphrases represent the most salient content units that form the summary’s building blocks. In abstractive approaches, keyphrases help identify central concepts that must be preserved in the reformulated summary. Thus, advances in speech KE directly contribute to improved summarization quality.

In recent years, significant progress has been observed in applying Large Language Models (LLMs) for abstractive summarization of spoken documents. Despite LLM successes in summarization tasks, the capabilities of small language models (sLLMs) are also actively being explored. Ryu et al. (2024) proposed KEITSum, a method that allows sLLMs to generate more relevant texts by extracting and emphasizing key elements (named entities and concluding sentences) in the source text. The authors emphasize that KEITSum enables LLMs to compete with proprietary LLMs in dialogue and news summarization tasks.

Furthermore, based on their operational principles, SSum models can be categorized as transcript-based (relying on speech recognition) and acoustic-based (utilizing acoustic information).

The former type of models primarily uses Automatic Speech Recognition (ASR) mechanisms to convert audio signals into text, and then applies text summarization methods to produce a summary (Kotey et al., 2023). Such methods are popular due to the significant progress made in text summarization (Wang, 2022). A typical implementation approach is a cascaded method, where a text summarization (TSum) model processes the input text transcribed by an ASR system (Matsuura et al., 2023).

Since the first type of SSum models relies on ASR, keyphrases in this case are also extracted from recognized text data. Consequently, their extraction is carried out using the same methods applicable to keyphrase extraction from written text.

A drawback of models using SSum on top of ASR results is their dependence on the availability and quality of ASR mechanisms, which may perform poorly, for example, in noisy environments or with low-resource languages (Wang, 2022). Additionally, this approach is limited by the lack of acoustic information that conveys the speaker's emotions (Matsuura et al., 2023).

An alternative direction is the development of the second type of models, which operate directly on the audio signal. However, approaches to speech summarization based on acoustic information are less popular and less studied in scientific literature.

One such approach recently proposed is E2E (end-to-end) SSum, which generates abstractive summaries directly from speech. It is not dependent on speech recognition errors and can fully utilize acoustic information. Despite promising results, such a model requires a large amount of training data and, in data-scarce conditions, generates unnatural sentences (Matsuura et al., 2023). The E2E approach has also been used in the Sen-SSum (sentence-wise speech summarization) algorithm, which generates text summaries from spoken documents on a sentence-by-sentence basis (Matsuura et al., 2024).

Shang et al. (2024) also proposed an E2E SSum model that uses LLMs to generate text summaries directly from speech features. The authors employ a multi-stage training approach, including ASR and text summarization (TSum) tasks as auxiliary ones, along with a curriculum learning strategy to facilitate the transition from TSum to SSum.

Wang (2022) proposed a three-stage extractive method, ESSumm, which also does not require speech recognition. The proposed architecture consists of segment generation, key segment extraction, and key segment merging stages. A pre-trained Wav2Vec2.0 model is used to extract deep features from the audio signal, which are then converted into multidimensional phoneme probability vectors using the k-means clustering algorithm. Subsequently, latent semantic analysis is applied to estimate the importance of audio segments based on the Euclidean distance between vectors. By directly processing the audio signal without ASR, the method is applicable to spontaneous conversations in live communication and potentially for presenting the final result as audio files rather than text.

Alongside the development of new models, research efforts are actively underway to improve existing approaches. For instance, Freisinger et al. (2025) presented a hierarchical thematic segmentation algorithm for transcripts that generates multi-level tables of contents. This approach is based on fine-tuning LLMs using an adaptive LoRA method and considers speech pauses. Kano et al. (2025) proposed an integrated approach to extractive and abstractive speech summarization, demonstrating improved results by using auxiliary information obtained from extractive summaries. BN et al. (2025) investigated the problem of hallucinations in LLMs when summarizing medical texts and proposed methods for their detection.

In this context, an important direction is the development of methods for automatically evaluating summary quality. Kroll and Kraus (2024) proposed a paradigm for evaluating LLM-based spoken document summarization systems, considering the specificities of AI-generated content and adapting social science methodologies to ensure the reliability and reproducibility of human evaluation. Gong et al. (2025) introduced the CREAM framework for evaluating meeting summaries without reference texts, based on logical reasoning and the matching of key facts.

In conclusion, the task of automatic keyword extraction from spoken language is primarily considered as a stage of speech summarization and, depending on the summarization model, can be implemented either by methods for extracting keyphrases from written text or by processing the acoustic signal.

## 6. Discussion

Automatic keyphrase extraction is a well-established subfield of Natural Language Processing with a rich history of methodological development. As this review outlines, the field has evolved from simple statistical measures (e.g., Luhn, 1957) and linguostatistical approaches (e.g., Rose et al., 2009) to graph-based algorithms (e.g., Mihalcea, 2004) and sophisticated machine learning models (e.g., Grootendorst, 2020). The primary goal has consistently been to automatically identify the most salient and topic-representative units within a document. Furthermore, the practice of automatic keyphrase annotation for corpus indexing, as detailed in Section 4, is a foundational practice in computational linguistics and information retrieval, providing the gold-standard data necessary for training and evaluating automated systems.

This review synthesizes this established knowledge to reveal a persistent and critical research gap: the significant disparity between KE methods developed for written text and those adapted for spoken language. While written-text KE is a mature field with numerous reliable methods, KE for speech remains a complex, underexplored challenge. This gap exists primarily because spoken language introduces variables largely absent in text: disfluencies, prosody, variable audio quality, and the inherent noise of the ASR process itself. Furthermore, the recent emergence of keyphrase generation using LLMs, as discussed in Section 3.3.5, represents a paradigm shift from extraction to creation, introducing new possibilities but also challenges related to factuality and controllability.

We aimed to provide a comprehensive analysis of contemporary theoretical approaches and practical solutions for automatic KE across both text and speech modalities. The methodology involved a systematic review and synthesis of literature spanning the terminological foundations, feature analysis, methodological classifications, and application domains of KE.

Our consolidation aligns with and extends previous surveys (e.g., Hasan and Ng, 2014). However, whereas previous reviews often focused on written text, our work explicitly contextualizes these methods against the challenges of spoken language and the new possibilities offered by LLMs. For instance, while traditional surveys discuss the superiority of ML methods like KEA (Sokolova and Mitrofanova, 2018) or KeyBERT for certain tasks, our review highlights that their effectiveness for speech is contingent on the accuracy of the upstream ASR system, a dependency that acoustic-direct methods seek to bypass.

Therefore, we posit that the field is transitioning from a focus on purely extractive methods for written text towards a more integrated and generative paradigm that must account for multimodality (text combined with acoustics) and leverage the generative power of LLMs. There is no one-size-fits-all solution; the optimal KE strategy is highly contingent on the input modality (text/audio), language resources, domain specificity, and the desired outcome (extraction or generation). For example, a user processing a large corpus of scientific articles in English might achieve the best results with a fine-tuned KeyBERT model or an LLM for generation. In contrast, analyzing a noisy podcast recording might benefit from a hybrid approach that first uses a robust ASR system and then applies a graph-based method such as TextRank, or alternatively, experiments with an emerging E2E model that incorporates acoustic cues.

This review has several inherent limitations. Firstly, as a synthetic overview, it does not include original empirical comparisons of the discussed algorithms. Consequently, claims about the relative performance of different methods are based on interpretations of the cited literature rather than on controlled experimental data. Secondly, the field of NLP, and particularly the subfield of LLMs, evolves at an extremely rapid pace. Therefore, this review provides a snapshot of the state-of-the-art as of early 2025, and newer architectures or findings may have emerged since the literature collection cutoff date. Finally, while we strove for comprehensiveness, the sheer volume of research may have led to the omission of some niche or very recent studies.

The generalizations and claims made, particularly regarding the performance and applicability of different method categories, are intended as conceptual guidelines. Their validity in any specific practical application depends on numerous local factors, such as the quality of the input data, the specific implementation of an algorithm, and the tuning of its parameters.

## 7. Conclusion

This systematic review has delineated the evolution of automatic keyphrase extraction from its foundational statistical methods to modern neural and generative paradigms. The analysis was guided by the research questions posed in the introduction, and our findings provide insights that address their core concerns.

Our review has explicitly detailed how different disciplines, from information retrieval and psycholinguistics to computational linguistics and speech processing, define, operationalize, and utilize keyphrases (RQ1). We have shown that these methodological contributions are not isolated; techniques from one domain often inform and enhance approaches in another (RQ2). For instance, psycholinguistic insights into prosodic cues directly contribute to developing more robust extraction pipelines for spoken language. Combining psycholinguistic theories (e.g., on the role of prosody and perception) with information retrieval metrics (e.g., TF-IDF, graph-based centrality) holds immense potential for creating more accurate and human-like extraction models, particularly for challenging domains like noisy speech or informal discourse (RQ3).

The limitations of current methods and language-specific challenges were a central focus of our analysis. We concluded that the most persistent limitations involve domain and genre adaptation, the processing of morphologically rich languages, and the significant gap in performance between written and spoken text processing. The scarcity of annotated resources for languages other than English remains a primary bottleneck (RQ5).

We find that LLMs are fundamentally reshaping the KE landscape, moving the field from extraction toward generation. While LLMs demonstrate a remarkable ability to generate absent, topic-related keyphrases, their application introduces new challenges related to output controllability, factual accuracy (hallucinations), and high computational cost (RQ6).

Finally, the most promising perspective for integration lies in multimodal systems that fuse textual features (from ASR transcripts) with acoustic-prosodic features (pitch, energy, duration) and leverage generative capabilities of LLMs. This integrated approach is essential for overcoming the current limitations and building robust, next-generation KE systems (RQ7).

The central conclusion that emerges is that KE is a profoundly context-dependent task. There is no universal solution. The future of effective KE lies not in a single algorithm but in developing a nuanced, multi-faceted toolkit. The optimal strategy must be carefully chosen based on the input modality, language, domain, and specific application requirements. Overcoming the current limitations, particularly concerning resource availability for non-English languages and the development of standardized multimodal frameworks, will be crucial for advancing the field and unlocking the full potential of automatic keyphrase extraction in managing the world's growing information ecosystem.

**References**

Abramov, E. G. (2011). Selection of keywords for a scientific article, *Nauchnaya periodika: problemy i resheniya*, 2, 35–40. *(In Russian)*

Abrosimov, K. I. and Mosyagina, A. G. (2022). Sodner for Russian nested named entity recognition, Computational Linguistics and Intellectual Technologies: Papers from the Annual International Conference "Dialogue", Moscow, Russia, June 15–18, 2022, 1–7. *(In English)*

Antipina, E. S. and Prokhorenkova, S. A. (2020). Modeling of creative linguistic personality on the material of romances on A. S. Pushkin's poem "Do not sing, beauty, in my presence…", *Prepodavatel XXI vek*, 2 (2). *(In Russian)*

Augenstein, I., Das, M., Riedel, S., Vikraman, L., and McCallum, A. (2017). SemEval 2017 Task 10: ScienceIE — Extracting Keyphrases and Relations from Scientific Publications. In Proceedings of the 11th International Workshop on Semantic Evaluation (SemEval–2017), 546–555, Vancouver, Canada. Association for Computational Linguistics. *(In English)*

BN, S., Shing, H.-C., Xu, L., Strong, M., Burnsky, J., Ofor, J., Mason, J.R., Chen, S., Srinivasan, S., Shivade, C., Moriarty, J., Cohen, J.P. (2025). Fact-Controlled Diagnosis of Hallucinations in Medical Text Summarization, Proceedings Interspeech 2025, Rotterdam, The Netherlands, August 17–21, 2025, 3070–3074. *(In English)*

Bolshakova, E. I., Klyshinskiy, E. S., Lande, D. V., Noskov, A. A., Peskova, O. V. and Yagunova, E. V. (2011). *Avtomaticheskaya obrabotka tekstov na estestvennom jazyke i kompyuternaya lingvistika* [Automatic processing of texts in natural language and computational linguistics: textbook], MIEM, Moscow, Russia. *(In Russian)*

Boudin, F. and Aizawa, A. (2025). An Analysis of Datasets, Metrics and Models in Keyphrase Generation. In Proceedings of the Fourth Workshop on Generation, Evaluation and Metrics (GEM²), 973–973, Vienna, Austria and virtual meeting. Association for Computational Linguistics. *(In English)*

Brezina, V., McEnery, T. and Wattam, S. (2015). Collocations in context: A new perspective on collocation networks, International Journal of Corpus Linguistics, 20 (2). *(In English)*

Campos, R., Mangaravite, V., Pasquali, A., Jatowt, A., Jorge, A., Nunes, C. and Jatowt, A. (2020). YAKE! Keyword Extraction from Single Documents using Multiple Local Features, Information Sciences Journal, 509, 257–289. *(In English)*

Chen, P. I. and Lin, S. J. (2010). Automatic keyword prediction using Google similarity distance, Expert Systems with Applications, 37(3), 1928–1938. *(In English)*

Chen, W., Chan, H. P., Li, P. and King, I. (2020). Exclusive Hierarchical Decoding for Deep Keyphrase Generation, Proceedings of the 58th Annual Meeting of the Association for Computational Linguistics. Association for

Computational Linguistics, Florence, Italy, July 5, 2020, 1095–1105. *(In English)*

Chen, Y.-N., Huang, Y., Lee, H.-Y. and Lee, L.-S. (2012). Unsupervised two-stage keyword extraction from spoken documents by topic coherence and support vector machine, IEEE International Conference on Acoustics, Speech and Signal Processing (ICASSP), Kyoto, Japan, March 25–30, 2012, 5041–5044. *(In English)*

Devlin, J., Chang, M.-W., Lee, K. and Toutanova, K. (2019). BERT: Pre-training of Deep Bidirectional Transformers for Language Understanding. *North American Chapter of the Association for Computational Linguistics*. *(In English)*

Dostal, M. (2011). Automatic Keyphrase Extraction Based on NLP and Statistical Methods, Proceedings of the Dateso 2011: Annual International Workshop on Databases, Texts, Specifications and Objects, Pisek, Czech Republic, April 20, 2011, 140–145. *(In English)*

Dubinina, E. Yu. (2020). Extracting keywords of a scientific article text in the process of creating an automatic abstract, *Vestnik VGU. Seriya: Filologiya. Zhurnalistika*, 1, 26–28. *(In Russian)*

Evert, S. (2022). Measuring keyness, Digital Humanities 2022: Conference Abstracts, Tokyo, Japan, 25–29 July 2022, 202–205. *(In Russian)*

Freisinger, S., Seeberger, P., Ranzenberger, T., Bocklet, T. and Riedhammer, K. (2025). Towards Multi-Level Transcript Segmentation: LoRA Fine-Tuning for Table-of-Contents Generation, Proceedings Interspeech 2025, Rotterdam, The Netherlands, August 17–21, 2025, 276–280. *(In Russian)*

Gabrielatos, C. (2018). Keyness Analysis: nature, metrics and techniques, Corpus Approaches to Discourse: A critical review, Routledge, Oxford. *(In English)*

Gallina, Y., Boudin, F. and Daille B. (2019). KPTimes: A Large-Scale Dataset for Keyphrase Generation on News Documents, Proceedings of the 12th International Conference on Natural Language Generation, Association for Computational Linguistics, Tokyo, Japan, 130–135. *(In English)*

Glazkova, A., Morozov, D. (2024). Exploring fine-tuned generative models for keyphrase selection: A case study for Russian. DAMDID-2024. https://doi.org/10.1007/978-3-032-03997-2_7 *(In English)*

Glazkova, A., Morozov, D., Garipov, T. (2025). Key algorithms for keyphrase generation: Instruction-based LLMs for Russian scientific keyphrases. Analysis of Images, Social Networks and Texts, Springer Nature Switzerland, Cham, 107–119. *(In English)*

Glazkova, A., Morozov, D., Mitrofanova, O., Savchuk, S. (2024). Generation of keywords for regional media texts using large language models [Generatsiya klyuchevyh slov dlja tekstov regionalnyh SMI s pomoshchyu bolshih yazykovyh modeley]. International scientific conference dedicated to the 20th anniversary of the Russian National Corpus, The V.V. Vinogradov Russian Language Institute of the Russian Academy of Sciences, Moscow, Russia, 37–39. *(In English)*

Gong, Z., Ai, L., Deshpande, H., Johnson, A., Phung, E., Wu, Z., Emami, A. and Hirschberg, J. (2025). Comparison-Based Automatic Evaluation for Meeting Summarization, Proceedings Interspeech 2025, Rotterdam, The Netherlands, August 17–21, 2025, 291–295. *(In English)*

Grineva, M. and Grinev, M. (2009). Analysis of text documents for extracting thematically grouped key terms, *Trudy Instituta sistemnogo programmirovaniya RAN*, 16, 155–165. *(In Russian)*

Grootendorst, M. (2020). KeyBERT: Minimal Keyword Extraction with BERT [Online], available at: http://doi.org/10.5281/zenodo.4461265 (Accessed 11.10.2025). *(In English)*

Grudeva, E. V. and Churilina, L. N. (2019). Retelling as a secondary text: linguistic and methodological potential, *Magnitogorskiy gosudarstvennyy tekhnicheskiy universitet im. G.I. Nosova*, Magnitogorsk, Russia. *(In Russian)*

Grudeva, E. V. and Gubushkina, A. A. (2020). Selection of keywords and oral retelling as secondary texts (based on the secondary speech activity of 6th grade students), *Vestnik Cherepovetskogo gosudarstvennogo universiteta*, 2 (95). *(In Russian)*

Gulyaev, O. V. and Lukashevich, N. V. (2013). Automatic classification of texts based on section heading, *Novyye informatsionnyye tekhnologii v avtomatizirovannykh sistemakh*, 16, 238–244. *(In Russian)*

Guseva, D., Mitrofanova, O. and Dolgushin, M. (2025). Human and Machine Keyphrase Perception in Russian Text and Speech, Speech and Computer: 26th International Conference, SPECOM 2024, Belgrade, Serbia, November 25–28, 2024, 265–280. *(In English)*

Hasan, K. and Ng, V. (2014). Automatic Keyphrase Extraction: A Survey of the State of the Art, Proceedings of the 52nd Annual Meeting of the Association for Computational Linguistics, June 23 –25, 2014, Baltimore, Maryland, USA, 1262–1273. *(In English)*

Hulth, A. (2003). Improved Automatic Keyword Extraction Given More Linguistic Knowledge. In Proceedings of the 2003 Conference on Empirical Methods in Natural Language Processing, 216–223. *(In English)*

Jacquemin, C. and Bourigault, D. (2003). Term extraction and automatic indexing, Handbook of Computational Linguistics, Oxford University Press, 599–615. *(In English)*

Jones, K. S. (1972). A statistical interpretation of term specificity and its application in retrieval, Journal of Documentation, 28 (1), 11–21. *(In English)*

Kamshilova, O. N. (2013). Small forms of scientific text: keywords and abstract (informational aspect), *Izvestiya Rossiyskogo Gosudarstvennogo pedagogicheskogo universiteta im. A.I. Gertsena*, 156, 106–117. *(In Russian)*

Kano, T., Ogawa, A., Delcroix, M., Fukuda, R., Chen, W., Watanabe, S. (2025). Pick and Summarize: Integrating Extractive and Abstractive Speech Summarization, Proceedings Interspeech 2025, Rotterdam, The Netherlands, August 17–21, 2025, 281–285. *(In English)*

Kilgarriff, A. (2009). Simple maths for keywords, Proceedings of the Corpus Linguistics 2009 Conference, Liverpool, UK, July 20–23, 2009, 1–6. *(In English)*

Kodzasov, S. V. and Krivnova, O. F. (2001). *Obshchaya fonetika* [General phonetics], Moscow, Russia. *(In Russian)*

Koloski, B., Pollak, S., Škrlj, B. and Martinc, M. (2021). Extending Neural Keyword Extraction with TF-IDF tagset matching, Proceedings of the EACL Hackashop on News Media Content Analysis and Automated Report Generation, April, 2021, 22–29. *(In English)*

Kotey, S., Dahyot, R. and Harte, N. (2023). Query Based Acoustic Summarization for Podcasts, Proc. Interspeech 2023, Dublin, Ireland, August 20–24, 2023, 1483–1487. *(In English)*

Krapivin, M., Autayeu, A., Marchese, M., Blanzieri, E., Segata, N. (2010). Keyphrases Extraction from Scientific Documents: Improving Machine Learning Approaches with Natural Language Processing. In: Chowdhury, G., Koo, C., Hunter, J. (eds) The Role of Digital Libraries in a Time of Global Change, ICADL 2010, Lecture Notes in Computer Science, vol. 6102, Springer, Berlin, Heidelberg. https://doi.org/10.1007/978-3-642-13654-2_12 *(In English)*

Krasavina, V. D. and Mirzagitova, A. R. (2015). Optimization of search in LeadScanner system using automatic extraction of keywords and word combinations, *Trudy mezhdunarodnoy konferentsii “Korpusnaya lingvistika–2015”*, St. Petersburg, Russia. *(In Russian)*

Kroll, M. and Kraus, K. (2024). Optimizing the role of human evaluation in LLM-based spoken document summarization systems, Proc. Interspeech 2024, Kos Island, Greece, September 1–5, 2024, 1935–1939. *(In English)*

Le-Duc, K., Nguyen, K.-N., Vo-Dang, L. and Hy, T.-S. (2024). ‘Real-time Speech Summarization for Medical Conversations’, Proc. Interspeech 2024, Kos Island, Greece, September 1–5, 2024, 1960–1964. *(In English)*

Lee, H.-y., Shiang, S.-R., Yeh, Ch.-F., Chen, Y.-N., Huang, Y., Kong, S.-y. and Lee, L.-S. (2014). Spoken Knowledge Organization by Semantic Structuring and a Prototype Course Lecture System for Personalized Learning, IEEE/ACM Transactions on Audio, Speech, and Language Processing, 22, 883–898. *(In English)*

Lee, W., Chun, M., Jeong, H., Jung, H. (2023). Toward keyword generation through large language models. Companion Proceedings of the 28th International Conference on Intelligent User Interfaces, 37–40. https://doi.org/10.1145/3581754.3584126 *(In English)*

Litvak, M. (2013). DegExt: A language-independent keyphrase extractor, Journal of Ambient Intelligence and Humanized Computing, 4, 377–387. *(In English)*

Luhn, H.P. (1957). A Statistical Approach to Mechanized Encoding and Searching of Literary Information, IBM Journal of Research and Development, 1 (4), 309–317. *(In English)*

Martínez-Cruz, R., López-López, A. J., and Portela, J. (2023). ChatGPT vs state-of-the-art models: a benchmarking study in keyphrase generation task. arXiv preprint, arXiv:2304.14177. *(In English)*

Martínez-Cruz, R., López-López, A. J., Portela, J. (2025) ChatGPT vs state-of-the-art models: Abenchmarking study in keyphrase generation task, Applied Intelligence, 55 (1), 50. *(In English)*

Marujo, L., Gershman, A., Carbonell, J., Frederking, R., and Neto, J. P. (2013). Supervised topical key phrase extraction of news stories using crowdsourcing, light filtering and co-reference normalization. arXiv preprint, arXiv:1306.4886. *(In English)*

Maskey, S. and Hirschberg, J. (2005). Comparing lexical, acoustic/prosodic, structural and discourse features for speech summarization, Proc. Interspeech 2005, Lisbon, Portugal, September 4 – 8, 2005, 621–624. *(In English)*

Matsuura, K., Ashihara, T., Moriya, T., Mimura, M., Kano, T., Ogawa, A. and Delcroix, M. (2024). Sentence-wise Speech Summarization: Task, Datasets, and End-to-End Modeling with LM Knowledge Distillation, Proc. Interspeech 2024, Kos Island, Greece, September 1–5, 2024, pp. 1945–1949. *(In English)*

Matsuura, K., Ashihara, T., Moriya, T., Tanaka, T., Kano, T., Ogawa, A. and Delcroix, M. (2023). Transfer Learning from Pre-trained Language Models Improves End-to-End Speech Summarization, Proc. Interspeech 2023, Dublin, Ireland, August 20–24, 2023, 2943–2947. *(In English)*

Meng, R., Zhao, S., Han, S., He, D., Brusilovsky, P. and Chi, Y. (2017) Deep Keyphrase Generation, Proceedings of the 55th Annual Meeting of the Association for Computational Linguistics (Volume 1: Long Papers), Association for Computational Linguistics, Vancouver, Canada, 582–592. *(In English)*

Meng, R., Zhao, S., Han, S., He, D., Brusilovsky, P., and Chi, Y. (2017). Deep Keyphrase Generation, in Proceedings of the 55th Annual Meeting of the Association for Computational Linguistics (Volume 1: Long Papers), 582–592, Vancouver, Canada. Association for Computational Linguistics. *(In English)*

Mihalcea, R. TextRank: Bringing order into texts, in Proc. EMNLP, 2004, 4. 404–411. *(In English)*

Mijić, J., Dalbelo Bašić, B., Šnajder, J. (2010). Robust Keyphrase Extraction for a Largescale Croatian News Production System, in Proceedings of the 2010 Conference on Empirical Methods in Natural Language Processing (EMNLP-2010), 59–99. *(In English)*

Mitrofanova, O. A. and Gavrilik, D. A. (2022). Experiments on automatic extraction of key expressions in stylistically diverse corpora of Russian texts, *Terra Linguistica*, 13 (4), 22–40. *(In Russian)*

Morozov, D. A., Glazkova, A. V., Tyutyul'nikov, M. A. and Iomdin, B. L. (2023). Generation of keywords for annotations of Russian scientific articles, *Vestnik NSU. Seriya: Lingvistika i mezhkul'turnaya kommunikatsiya*, 1. *(In Russian)*

Moskvina, A. D., Erofeeva, A. R., Mitrofanova, O. A. and Kharabet, Ya. K. (2017). Automatic extraction of keywords and word combinations from Russian corpus of texts using RAKE algorithm, *Trudy Mezhdunarodnoy konferentsii "Korpusnaya lingvistika–2017" (Sankt-Peterburg, 27–30 iyunya 2017 g.)*, Izd-vo SPbGU, Russia, 268–275. *(In Russian)*

Moskvina, A., Sokolova, E. and Mitrofanova, O. (2018). KeyPhrase extraction from the Russian corpus on Linguistics by means of KEA and RAKE algorithm, Data Analytics and Management in Data Intensive Domains: XX International Conference DAMDID/RCDL'2018: Conference Proceedings, Moscow, Russia, October 9–12, 2018, 369–372. *(In English)*

Moskvitina, T. N. (2009). Keywords and their functions in scientific text, *Vestnik ChGPU*, 11, 270–283. *(In Russian)*

Moskvitina, T. N. (2018). Methods for extracting keywords when abstracting a scientific text, *Vestnik Tomskogo gosudarstvennogo universiteta*, 8, 45–50. *(In Russian)*

Nguyen, T. D., Kan, M-Y. (2007). Keyphrase Extraction in Scientific Publications. In: Goh, D. H-L., Cao, T. H., Sølvberg, I. T., Rasmussen, E. (eds) Asian Digital Libraries. Looking Back 10 Years and Forging New Frontiers, ICADL 2007, Lecture Notes in Computer Science, vol. 4822, Springer, Berlin, Heidelberg. P. 317–326. https://doi.org/10.1007/978-3-540-77094-7_41 *(In English)*

Papusha, I. S. (2008). Complex syntactic whole: keywords or herms, *Vestnik Assotsiatsii VUZov turizma i servisa*, 3, 48–54. *(In Russian)*

Piotrovskiy, R. G., Bektaev, K. B. and Piotrovskaya, A. A. (1977). *Matematicheskaya lingvistika: ucheb. posobiye dlya ped. institutov* [Mathematical linguistics: textbook for pedagogical institutes], Vysshaya shkola, Moscow, Russia. *(In Russian)*

Popova, S. V. and Danilova, V. V. (2014). 'Representation of documents in the task of clustering scientific text annotations', *Nauchno-tekhnicheskiy vestnik informatsionnykh tekhnologiy, mekhaniki i optiki*, 1 (89), 99–107. *(In Russian)*

Riedhammer, K., Favre, B. and Hakkani-Tur, D. (2010). Long story short–global unsupervised models for keyphrase based meeting summarization, Speech Communication, 52 (10), 801–815. *(In English)*

Rose, S. J., Cowley, W. E., Crow, V. L. and Cramer, N. O. (2009). Rapid Automatic Keyword

Extraction for Information Retrieval and Analysis, Text Mining: Applications and Theory, 1–20. *(In English)*

Ryu, S., Do, H., Kim, Y., Lee, G.G. and Ok, J. (2024). Key-Element-Informed sLLM Tuning for Document Summarization, Proc. Interspeech 2024, Kos Island, Greece, September 1–5, 2024, 1940-1944. *(In English)*

Sakharnyy, L. V. (1982). Actual division and text compression (on the use of informatics methods in psycholinguistics, *Teoreticheskiye aspekty derivatsii*, Perm, Russia. *(In Russian)*

Sakharnyy, L. V. and Shtern, A. S. (1988). Selection of keywords as a type of text, *Leksicheskiye aspekty v sisteme professional'no-orientirovannogo obucheniya inoyazychnoy rechevoy deyatel'nosti*, Perm, Russia, 34–51. *(In Russian)*

Shang, H., Li, Z., Guo, J., Li, S., Rao, Z., Luo, Y., Wei, D. and Yang, H. (2024). An End-to-End Speech Summarization Using Large Language Model, Proc. Interspeech 2024, Kos Island, Greece, September 1–5, 2024, 1950–1954. *(In English)*

Shao, L., Zhang, L., Peng, M., Ma, G., Yue, H., Sun, M., and Su, J. (2024). One2set+ large language model: Best partners for keyphrase generation, in Proceedings of the 2024 Conference on Empirical Methods in Natural Language Processing, 11140–11153. *(In English)*

Shekhtman, N. A. (2005). *Ponimaniye rechevogo proizvedeniya i gipertekst* [Understanding of a speech work and hypertext], Izd-vo OGPU, Orenburg, Russia. *(In Russian)*

Sheremetyeva, S. O. and Osminin, P. G. (2015). Methods and models of automatic keyword extraction, *Vestnik YUrGU. Seriya "Lingvistika"*, 12 (1), 76–81. *(In Russian)*

Sokolova, E. V. and Mitrofanova, O. A. (2018). Automatic extraction of keywords and word combinations from Russian texts using KEA algorithm, *Kompyuternaya lingvistika i vychislitel'nyye ontologii*, 157–165. *(In Russian)*

Song, M., Geng, X., Yao, S., Lu, S., Feng, Y., Jing, L. (2023). Large language models as zero-shot keyphrase extractor: A preliminary empirical study. arXiv preprint arXiv:2312.15156 *(In English)*

Song, M., Jiang, H., Shi, S., Yao, S., Lu, S., Feng, Y., Liu, H., and Jing, L. (2023). Is ChatGPT a good keyphrase generator? A preliminary study. arXiv preprint, arXiv:2303.13001. *(In English)*

Sterckx, L., Demeester, T., Deleu, J., et al. (2018). Creation and evaluation of large keyphrase extraction collections with multiple opinions. *Language Resources & Evaluation*, 52, 503–532. *(In English)*

Su, J., Zhang, L., Hassanzadeh, H. R. and Schaaf, T. (2022). Extract and Abstract with BART for Clinical Notes from Doctor-Patient Conversations, Proc. Interspeech 2022, Incheon, Korea, September 18–22, 2022, 2488–2492. *(In English)*

Svetozarova, N. D. and Shtern, A. S. (1989). Key and phonematically highlighted words of the text, *Eksperimentalnaya fonetika*, Moscow, Russia, 157–170. *(In Russian)*

Troshina, A. (2025). Text Preprocessing for Keyword and Key Phrase Extraction. In: Bakaev, M., et al. Internet and Modern Society. Human-Computer Communication. IMS 2024. Communications in Computer and Information Science, vol 2534. Springer, Cham, 105–112. https://doi.org/10.1007/978-3-031-96177-9_9 *(In English)*

Tsarfaty, R., Seddah, D., Kübler, S. and Nivre, J. (2013). Parsing Morphologically Rich Languages: Introduction to the Special Issue, Computational Linguistics, 3 9(1), 15–22. *(In English)*

Umair, M., Sultana, T., and Lee, Y.-K. (2024). Pre-trained language models for keyphrase prediction: A review. *ICT Express*, 10(4), 871–890. *(In English)*

Umair, M., Sultana, T., Lee, Y. K. (2024) Pre-trained language models for keyphrase prediction: A review. ICT Express, 10 (4), 871–890. https://doi.org/10.1016/j.icte.2024.05.015 *(In English)*

Vanyushkin, A. S. and Grashchenko, L. A. (2016). Methods and algorithms for extracting keywords, *Novyye informatsionnyye tekhnologii v avtomatizirovannykh sistemakh*, 19, 85–93. *(In Russian)*

Vanyushkin, A. S. and Grashchenko, L. A. (2017). Evaluation of keyword extraction algorithms: tools and resources, *Novyye informatsionnyye tekhnologii v avtomatizirovannykh sistemakh*, 20. *(In Russian)*

Vanyushkin, A. S. and Grashchenko, L. A. (2018). On the marking of text corpora with keywords, *Novyye informatsionnyye tekhnologii v avtomatizirovannykh sistemakh*, 21, 207–211.

Vartakavi, A. and Garg, A. (2020). Podsumm: Podcast audio summarization [Online], available at: https://arxiv.org/pdf/2009.10315 (Accessed 3.11.2025). *(In English)*

Vasileva, V. V. and Kon'kov, V. I. (2015). *Ustnaya rech: praktikum* [Spoken speech:

workshop], S.-Peterb. gos. un-t, St. Petersburg, Russia. *(In Russian)*

Vinogradova, N. V. and Ivanov, V. K. (2016). ‘Modern methods of automated extraction of keywords from text’, *Informatsionnyye resursy Rossii*, 4, 13–18. *(In Russian)*

Wan, X. and Xiao, J. (2008). Single document keyphrase extraction using neighborhood knowledge. In Proceedings of the 23rd National Conference on Artificial Intelligence, volume 2, AAAI Press, 855–860. *(In English)*

Wang, J. (2022). ESSumm: Extractive Speech Summarization from Untranscribed Meeting, Proc. Interspeech 2022, Incheon, Korea, September 18–22, 2022, 3243–3247. *(In English)*

Wang, S., Dai, S., and Jiang, J. (2024). Thinking like an author: A zero-shot learning approach to key phrase generation with large language model. In A. Bifet, J. Davis, T. Krilavičius, M. Kull, E. Ntoutsi, and I. Žliobaitė (Eds.), Machine Learning and Knowledge Discovery in Databases. Research Track. Springer Nature Switzerland, Cham, 335–350. *(In English)*

Wienecke, Y. (2020). Automatic Keyphrase Extraction from Russian-Language Scholarly Papers in Computational Linguistics, University Honors Theses, Portland State University. *(In English)*

Wilson, A. (2013). Embracing Bayes factors for key item analysis in corpus linguistics, New approaches to the study of linguistic variability. Language Competence and Language Awareness in Europe, 4, 3–11. *(In English)*

Xiong, L., Chuan Hu, Chenyan Xiong, Campos D., and Overwijk, A. (2019). Open Domain Web Keyphrase Extraction Beyond Language Modeling. In Proceedings of the 2019 Conference on Empirical Methods in Natural Language Processing and the 9th International Joint Conference on Natural Language Processing (EMNLP-IJCNLP), pages 5175–5184, Hong Kong, China. Association for Computational Linguistics. *(In English)*

Xiong, L., Hu, C., Xiong, C., Campos, D. and Overwijk, A. (2019). Open Domain Web Keyphrase Extraction Beyond Language Modeling. In Proceedings of the 2019 Conference on Empirical Methods in Natural Language Processing and the 9th International Joint Conference on Natural Language Processing (EMNLP-IJCNLP), Association for Computational Linguistics, Hong Kong, China, 5175–5184. *(In English)*

Yagunova, E. V. (2004). The role of keywords in the perception of spoken and written text (based on the Russian language), *Chelovek pishushchiy i chitayushchiy: problemy i nablyudeniya: Materialy i nablyudeniya: Materialy mezhdunarodnoy konferentsii 14-16 marta 2002 g. Sankt-Peterburg*, Izd-vo SPbGU, Russia, 197–204. *(In Russian)*

Yagunova, E. V. (2010). ‘Experiment and calculations in the analysis of keywords of a fictional text’, *Filosofiya yazyka. Lingvistika. Lingvodidaktika*, 1, 83–89.

Zakharov, V. P. and Khokhlova, M. V. (2014). ‘Extraction of terminological phrases from special texts based on various association measures’, *XVII Vserossiyskaya obyedinennaya konferentsiya “Internet i sovremennoye obshchestvo” (IMS-2014)*, St. Petersburg, Russia. *(In Russian)*

Zhang, C., Wang, H., Liu, Y., Wu, D., Liao, Y. and Wang, B. (2008). Automatic keyword extraction from documents using conditional random fields, Journal of Computational Information Systems, 4(3), 1169–1180. *(In English)*

***Conflicts of interest: the authors have no conflict of interest to declare.***

***Конфликты интересов: у авторов нет конфликта интересов для декларации.***

***The authors have read and approved the final manuscript.***

***Авторы прочитали и одобрили окончательный вариант рукописи.***

**Daria D. Guseva**, teaching assistant at the Department of Phonetics and Language Teaching Methods, Faculty of Philology, Saint-Petersburg State University, Saint-Petersburg, Russia.

**Дарья Дмитриевна Гусева**, ассистент кафедры фонетики и методики преподавания иностранных языков, филологический факультет, Санкт-Петербургский государственный университет, Санкт-Петербург, Россия.

**Olga A. Mitrofanova**, PhD in Linguistics, Associate Professor, Department of Mathematical Linguistics, Faculty of Philology, Saint-Petersburg State University, Saint-Petersburg, Russia.

**Ольга Александровна Митрофанова**, доцент, кандидат филологических наук, доцент кафедры математической лингвистики, Санкт-Петербургский государственный университет, Санкт-Петербург, Россия.